\documentclass[letterpaper]{article} 
\usepackage{aaai2027}
\nocopyright
\usepackage[hyphens]{url}  
\usepackage{graphicx} 
\def\UrlFont{\rm}  
\usepackage{natbib}  
\usepackage{caption} 
\usepackage{booktabs}
\usepackage{amsmath,amssymb}
\newcommand{\method}{Robo-Dopamine 2.0}

\title{\method{}: History-Conditioned and OOD-Aware Process Reward Modeling for Robotic Manipulation}
\author{
Yijie Xu\textsuperscript{\rm 1,\rm 2}\equalcontrib,
Haopeng Jin\textsuperscript{\rm 3,\rm 4}\equalcontrib,
Run Zhou\textsuperscript{\rm 5}\equalcontrib,
Shengbang Liu\textsuperscript{\rm 6}\equalcontrib,\\
Sixiang Chen\textsuperscript{\rm 1,\rm 2},
Hongyang Cheng\textsuperscript{\rm 2},
Sicheng Hu\textsuperscript{\rm 1,\rm 2},
Peterson Co\textsuperscript{\rm 1,\rm 2},\\
Jinwen Luo\textsuperscript{\rm 4},
Huajie Tan\textsuperscript{\rm 1,\rm 2},
Shanghang Zhang\textsuperscript{\rm 1}\corresponding
}
\affiliations{
\textsuperscript{\rm 1}State Key Laboratory of Multimedia Information Processing, School of Computer Science, Peking University\\
\textsuperscript{\rm 2}EvoPhys AI\\
\textsuperscript{\rm 3}Beijing University of Posts and Telecommunications\\
\textsuperscript{\rm 4}Tencent\\
\textsuperscript{\rm 5}Renmin University of China\\
\textsuperscript{\rm 6}Sun Yat-sen University
}

\begin{document}

\maketitle

\begin{abstract}
Vision-language-action (VLA) models have improved robotic manipulation, yet
their policies remain vulnerable to compounding errors, unseen scene changes,
and off-trajectory states. Reinforcement learning (RL) offers a promising
route to refine pretrained VLA policies, but its effectiveness depends on
informative rewards. Sparse success signals make exploration inefficient,
whereas manually engineered dense rewards are costly, brittle, and
task-specific. Learned visual reward models provide dense feedback, but
existing robotic reward models often rely on static before--after observations
and struggle to capture execution context and failure semantics. This leads to
\emph{temporal ambiguity}, where similar observations correspond to different
latent progress due to hidden contact events, repeated phases, or reordered
subgoals, and limits their ability to distinguish robustness-preserving
variations from task-invalid failures under out-of-distribution (OOD)
execution. We introduce \textbf{\textit{Robo-Dopamine 2.0}}, a history- and OOD-aware process
reward model that retains a pairwise prediction interface.
\method{} improves progress estimation under these conditions through two
complementary designs: (1) a history-conditioned pairwise reward formulation
that uses source-aligned reference panels for synthetic OOD queries and
observed rollout history for online queries, while preserving the actual queried
endpoints;
and (2) an OOD-aware signed progress space that organizes positive,
robustness-preserving, and negative states into a unified progress
representation. To improve optimization of fine-grained progress ordering, we
further develop a \textbf{\textit{Signed-Hop Curriculum}} with transition-aware replay
that progressively learns coarse execution ordering before fine-grained
progress calibration. We also construct an OOD trajectory dataset and a
five-family benchmark for comprehensive evaluation. Reference panels improve mean visual order consistency (VOC) from 0.967 to
0.986 and OOD-robust VOC from 0.906 to 0.958. Under the same 400K pairwise-reward budget, Signed-Hop training with 25\%
replay reaches 0.9872 mean VOC, compared with 0.9858 for a matched-pool
shuffled control. In downstream RL, the full model reaches 86.8\% mean
RoboTwin success and 71/80 successful real-world insertions.
\end{abstract}

\section{Introduction}

Large-scale imitation learning and vision-language-action models have
substantially improved visuomotor policy generalization from diverse
demonstrations~\citep{RT12022,RT22023,OpenVLA2024}.
However, long-horizon, contact-rich manipulation remains challenging due to
compounding errors, unseen scene changes, and off-trajectory states that
require recovery.
Reinforcement learning can improve policies through interaction, but its
practical deployment on real robots critically depends on rewards that are
dense, task-relevant, and reliable beyond the demonstration distribution.
Sparse terminal rewards make exploration inefficient, whereas handcrafted
dense rewards are brittle, task-specific, and expensive to
engineer~\citep{PEBBLE2021,RLVLMF2024,RoboCLIP2023}.

Recent visual and language-conditioned reward models address this bottleneck by
learning dense feedback from observations and task descriptions
~\citep{VIP2023,LIV2023,VLMRewards2023,RLVLMF2024}; robot-specific methods
further introduce stage- or process-aware progress estimation
~\citep{VICtoR2024,RoboDopamine2025,SARM2025,VLAC2025}.
Nevertheless, existing approaches remain limited in accurately evaluating
execution progress under challenging manipulation scenarios.
\textbf{\textit{First}}, static endpoint pairs suffer from \emph{temporal aliasing}: visually
similar states may correspond to different latent progress due to hidden
contact events, repeated phases, or reordered subgoals
~\citep{Ni2022RecurrentPOMDP,Guhur2023HistoryAware,Fu2024TemporalOT}.
\textbf{\textit{Second}}, robust execution requires distinguishing semantics-preserving
variations from true failures: occlusions or distractors may preserve task
progress, whereas failed grasps, wrong-object interactions, and incomplete
releases should receive negative feedback. However, successful
in-distribution trajectories provide limited supervision for learning such
failure-aware distinctions.

To address these challenges, we introduce \textbf{\textit{Robo-Dopamine 2.0}}, a
history- and OOD-aware pairwise reward learning framework that retains the
pairwise progress formulation used in prior process reward
modeling~\citep{RoboDopamine2025}. \method{} improves phase-aware and
failure-aware progress estimation through two complementary designs:
\textbf{\textit{(1) a history-conditioned pairwise reward formulation}} that
uses same-episode expert history for standard queries and observed rollout
history for online queries, while synthetic OOD queries use phase-aligned
successful-reference panels without replacing their edited endpoints; and \textbf{\textit{(2) an OOD-aware signed progress space}} that
organizes execution states into positive, robustness-preserving, and negative
branches, enabling the model to represent valid progress,
semantics-preserving variations, failures, and recovery within a unified
pairwise training interface.

Furthermore, we introduce a \textbf{\textit{Signed-Hop Curriculum}} with
transition-aware replay to progressively learn execution ordering.
The curriculum first captures coarse-grained ordering between successful,
robust, and failed states, and then refines fine-grained progress differences.
The replay strategy preserves informative large-transition pairs during
training, improving progress calibration under diverse execution scenarios.
We evaluate \method{} through a trajectory-level visual order consistency (VOC)
evaluation across five benchmark families covering in-distribution, temporal,
and OOD scenarios.

Our contributions are summarized as follows:
\begin{itemize}
    \item We propose \textbf{\textit{Robo-Dopamine 2.0}}, a history- and OOD-aware pairwise reward framework that supports source-reference context for synthetic OOD queries and observed rollout context for online reward inference, together with a signed progress space for valid progress, robustness, failure, and recovery.
    \item We develop a \textbf{\textit{Signed-Hop Curriculum}} with transition-aware replay that
    improves learning of global execution ordering and fine-grained progress
    calibration.
    \item We present a trajectory-level visual order consistency (VOC)
    evaluation across five benchmark families covering in-distribution, temporal,
    and OOD scenarios, including controlled studies of static versus
    history-conditioned reward modeling.
\end{itemize}

\begin{figure*}[t]
\centering
\includegraphics[width=1\textwidth]{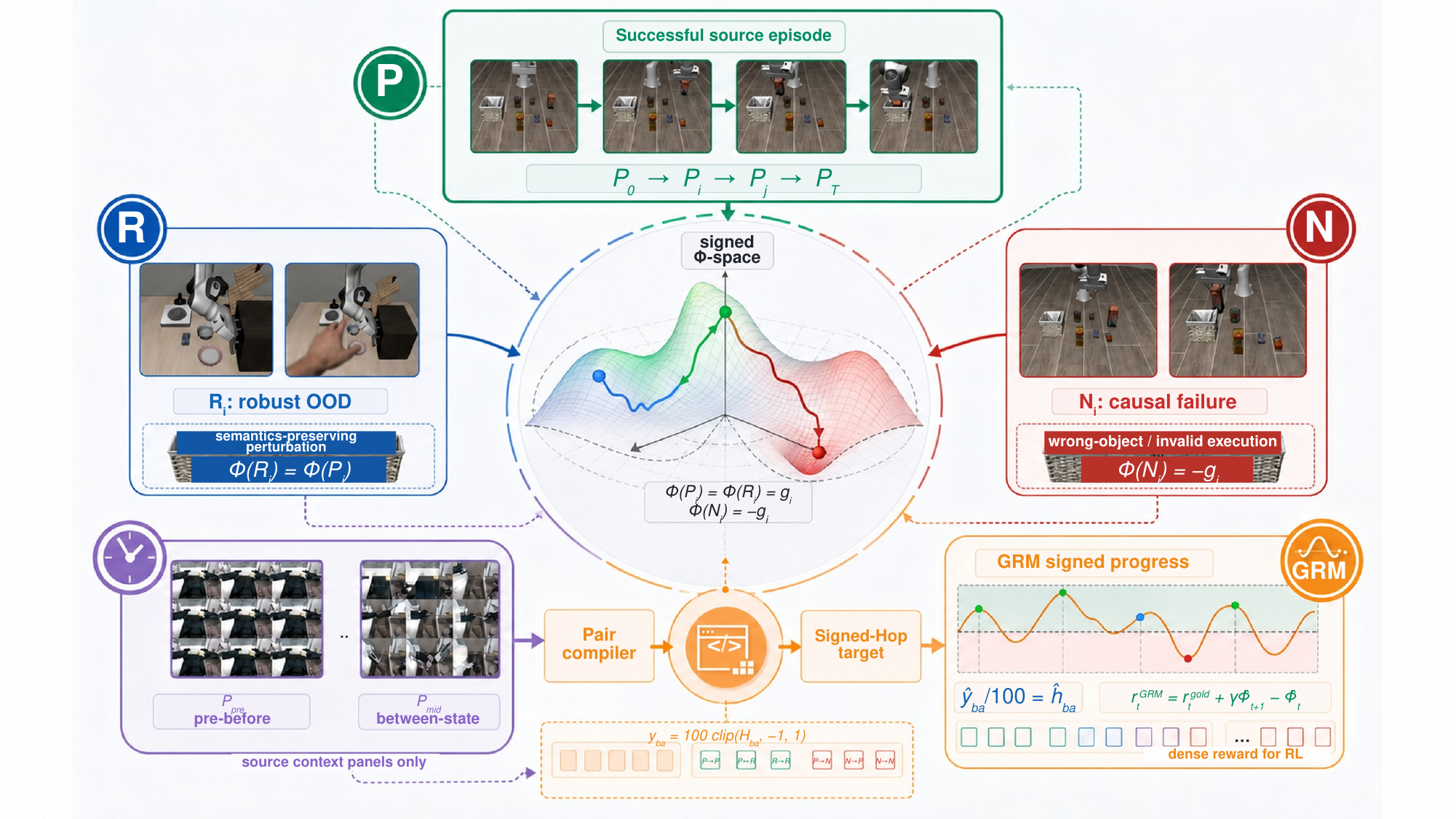}
\caption{Overview of \method{}. Standard queries use same-episode expert
history, online queries use the observed rollout history, and synthetic OOD
queries use source-aligned successful-reference panels while preserving their
edited endpoints. All context modes share the same pairwise GRM interface.}
\label{fig:method_overview}
\end{figure*}

\section{Related Work}
\label{sec:related_work}

\textbf{Generalist Robot Policies.}
Generalist robot policies and vision-language-action models scale visuomotor
learning and semantic generalization across tasks and embodiments. RT-1 studies
large-scale real-world robot control, RT-2 transfers web-scale semantic
knowledge into action prediction, and OpenVLA provides an open model for
adaptation to new manipulation domains~\citep{RT12022,RT22023,OpenVLA2024}.
These works establish the foundation for general robot policies, but do not
address process reward modeling for off-trajectory recovery. \method{} instead
focuses on the reward layer for refining policies under long-horizon and
out-of-distribution execution.

\vspace{0.5em}

\noindent
\textbf{Visual Reward Learning.}
Reward learning explores preference supervision~\citep{PEBBLE2021},
language-conditioned state changes~\citep{LOReL2022}, and feedback derived from
visual or video-language representations~\citep{RoboCLIP2023,VIP2023,LIV2023,
VideoLanguageCritic2024}. Recent foundation-model approaches learn pairwise
visual preferences, general-purpose rewards, or failure-aware feedback
~\citep{RLVLMF2024,VLMRewards2023,Adapt2Reward2024}, while VICtoR introduces
stage-aware vision-instruction rewards~\citep{VICtoR2024}. Despite these
advances, existing methods remain largely endpoint-oriented or success-centric
and do not combine query-dependent temporal context with explicit OOD
failure/recovery supervision. \method{} complements these approaches by jointly
modeling temporal-context-conditioned pairs and signed OOD progress within a
unified reward formulation.

\vspace{0.5em}

\noindent
\textbf{Temporal and Process-aware Reward Modeling.}
Temporal context has been explored in recurrent RL, history-aware manipulation
policies, and order-sensitive proxy rewards
~\citep{Ni2022RecurrentPOMDP,Guhur2023HistoryAware,Fu2024TemporalOT}.
These methods primarily introduce memory into policies or trajectory matching,
whereas \method{} targets temporal ambiguity within pairwise reward estimation.
Meanwhile, robotic process reward models investigate multi-view relative
progress, large-scale reward data, stage-aware estimation, and critic-like
architectures~\citep{RoboDopamine2025,RoboReward2026,VLAC2025,SARM2025,ARM2026};
failure- and stage-reasoning models provide complementary diagnostics
~\citep{AHA2024,MARVL2026}. \method{} bridges these directions by using
query-dependent temporal context---same-episode expert history for standard
queries, observed rollout history for online queries, and source-aligned
successful-reference panels for synthetic OOD queries---and by unifying semantics-preserving variations, failures, and
recovery within a signed pairwise progress space.

\section{Method}
\label{sec:method}

Figure~\ref{fig:method_overview} summarizes \method{}: ordered temporal context
addresses ambiguous endpoints, signed OOD supervision models valid progress,
failure, and recovery, and a Signed-Hop curriculum improves global-to-local
ordering. Closed-loop inference converts pairwise predictions into an online
potential for reinforcement learning.

\subsection{History-Conditioned Pairwise Reward Formulation}
\label{sec:method_history_formulation}

\method{} preserves the pairwise interface of prior process reward
modeling~\citep{RoboDopamine2025}. Given an instruction $c$, task-reference
anchors $\mathcal R=(r_s,r_g)$, ordered temporal context
$\mathcal C_{ba}$, and an ordered pair $(X_b,X_a)$, the General Reward
Model (GRM) predicts signed relative progress:
\begin{equation}
    \hat y_{ba}
    =
    f_\theta(c,\mathcal R,\mathcal C_{ba},X_b,X_a).
\end{equation}
Here $X_b$ and $X_a$ are the actual queried multi-view states. The reference
anchors specify the task start and goal but are not treated as observed
execution history. The static baseline sets
$\mathcal C_{ba}=\varnothing$.

Let $X_b=(b^h,b^l,b^r)$ and $X_a=(a^h,a^l,a^r)$, where $h,l,r$ denote the
canonical head/third-person, left-wrist, and right-wrist view slots. For a pair with temporal indices $t_b$ and $t_a$, we define
\begin{equation}
\begin{aligned}
\mathcal W_{\mathrm{pre}}
    &=\{t:0\leq t<\min(t_b,t_a)\},\\
\mathcal W_{\mathrm{mid}}
    &=\{t:\min(t_b,t_a)<t<\max(t_b,t_a)\}.
\end{aligned}
\end{equation}
For standard and online queries, these are indices in the queried episode or
rollout; for synthetic OOD pairs, they are indices in the aligned source
episode. Pair order determines the target direction, while panel frames remain
chronological. Each window is uniformly sampled into panels
$M_{\mathrm{pre}}^{1:K}$ and $M_{\mathrm{mid}}^{1:L}$. The two input
layouts are
\begin{equation}
\begin{aligned}
\mathcal I_{\mathrm{static}}
    &= [r_s,r_g,X_b,X_a],\\
\mathcal I_{\mathrm{ref}}
    &= [r_s,r_g,M_{\mathrm{pre}}^{1:K},X_b,\\
    &\qquad M_{\mathrm{mid}}^{1:L},X_a].
\end{aligned}
\end{equation}
Frames in each panel are ordered left-to-right and top-to-bottom. Empty cells
use a fixed padding image, and the prompt records the actual panel count so
that image and text positions remain aligned. Panels may be rendered offline
or assembled and cached by the data loader under the same ordering and padding
contract. For same-episode queries, pre-query and between-state panels summarize the
actual events preceding and separating the endpoints. For synthetic OOD pairs,
they instead provide nominal successful progression at the aligned source
indices.

Context provenance follows the query type. Standard training pairs use
temporal panels from the same expert episode as the queried states. Synthetic
OOD pairs use panels exclusively from the aligned successful source episode,
while $X_b$ and $X_a$ remain the edited observations; we refer to these as
\emph{source-only reference panels}. Online inference instead constructs panels from the observed rollout history,
as described in the \emph{Closed-Loop Reward Framework} subsection.

\subsection{Structured OOD Data Construction}
\label{sec:method_ood_construction}

We use OOD to denote counterfactual states outside the distribution of
successful source executions. Successful trajectories are drawn from the
RoboCasa~\citep{RoboCasa2024} and LIBERO~\citep{LIBERO2023} simulation
benchmarks and from real-robot AgiBot World
trajectories~\citep{AgiBotWorld2025}. Dataset-specific camera streams are
mapped to common view slots; unavailable auxiliary views are padded and masked
according to their availability. Data splits are formed at the source-episode
level before OOD generation, and every edited descendant remains in the same
split as its source episode.

Given a successful source trajectory $\tau$ and instruction $c$, we apply a
task- and phase-aware edit operator to construct an aligned counterfactual
trajectory:
\begin{equation}
\begin{aligned}
    \tilde{\tau}_z
        &= \mathcal E_z(\tau,c),\\
    z
        &\in
        \mathcal Z_{\mathrm{rob}}
        \cup
        \mathcal Z_{\mathrm{neg}}.
\end{aligned}
\end{equation}
Here $z$ denotes the edit type, $\mathcal E_z$ applies it over a phase-valid
interval, and $\tilde{\tau}_z$ contains source-index-aligned edited endpoints.
The sets $\mathcal Z_{\mathrm{rob}}$ and $\mathcal Z_{\mathrm{neg}}$ contain
semantics-preserving and task-invalid operators, respectively.

Negative operators include wrong-object interaction, empty grasp, non-release,
and target replacement, whereas robust operators include partial occlusion,
background variation, irrelevant distractors, and non-target color changes.
Operator availability is primitive-dependent, and each edited sequence uses one
operator specification over a phase-valid interval and across all available
camera views. Each edited frame retains its source episode,
frame index, view, and operator metadata for aligned pair compilation. Full
prompt, temporal-localization, and view-specific editing details are provided
in the supplementary.

For synthetic OOD pairs, the task-reference anchors and all temporal panels are
drawn from the aligned successful source episode, while the intervention
affects only the queried endpoints. Edited observations never enter these
source-reference panels. Online deployment follows the rollout-history construction described below and
therefore does not require source selection or source-index alignment.

\subsection{OOD-aware Signed Progress Space }
\label{sec:method_progress_space}
\label{sec:method_signed_hop}

For a successful source trajectory
$\tau=\{s_0,\ldots,s_{T-1}\}$, define
\begin{equation}
    g_i
    =
    \frac{i}{T-1}.
\end{equation}
Here $P_i$ and $R_i$ denote the original positive state and its aligned
semantics-preserving robust state, respectively; $N_i$ denotes an aligned
task-invalid state instantiated only at phase-valid indices $i>0$. For each
instantiated state, we assign
\begin{equation}
    \Phi(P_i)
    =
    \Phi(R_i)
    =
    g_i,
    \qquad
    \Phi(N_i)
    =
    -g_i.
\end{equation}
The sign separates valid execution from task-invalid failure, while
$|\Phi(N_i)|=g_i$ indexes failure depth by the aligned source phase.
Thus, $-g_i$ is a designed supervision coordinate rather than a calibrated
failure probability or a literal percentage of negative completion. Since
every instantiated negative state satisfies $i>0$, we have $\Phi(N_i)<0$;
such states represent task-invalid execution rather than arbitrary difficulty
or visual novelty.

Define $b(P_i)=b(R_i)=0$ and $b(N_i)=1$, where $b(\cdot)$ is the binary
negative-branch indicator. For a pair $(X_b,X_a)$, let
\begin{equation}
\begin{aligned}
    \phi_b
        &= \Phi(X_b),
    & \phi_a
        &= \Phi(X_a),\\
    \Delta_{ba}
        &= \phi_a-\phi_b,
    & L_{ba}
        &= -\max\{b(X_b),b(X_a)\}.
\end{aligned}
\end{equation}
Here $L_{ba}$ selects the lower endpoint of the normalization interval:
$L_{ba}=0$ for valid-only pairs and $L_{ba}=-1$ whenever either endpoint
belongs to the negative branch. The signed Hop target is
\begin{equation}
\mathcal H_{ba}
=
\begin{cases}
0,
    & \Delta_{ba}=0,\\[2pt]
\dfrac{\Delta_{ba}}{1-\phi_b},
    & \Delta_{ba}>0,\\[6pt]
\dfrac{\Delta_{ba}}{\phi_b-L_{ba}},
    & \Delta_{ba}<0,
\end{cases}
\label{eq:signed_hop}
\end{equation}
The corresponding regression label is
\begin{equation}
    y_{ba}
    =
    100\,
    \operatorname{clip}
    (\mathcal H_{ba},-1,1),
\end{equation}
where the factor $100$ preserves the output scale of the original GRM
interface; it does not turn the continuous Hop into a three-class target. Zero
denotes same-progress invariance rather than uncertainty or a generic neutral
class.
Following the Hop normalization of \citet{RoboDopamine2025},
$\mathcal H_{ba}$ measures the fraction of the available potential range in the
queried direction: positive transitions are normalized toward the goal $1$,
whereas negative transitions are normalized toward the branch floor $L_{ba}$.
Replacing the original floor $0$ with the pair-dependent $L_{ba}$ extends the
same invertible mapping to failure and recovery; clipping is used only as a
numerical safeguard.

Training includes valid pairs $U_i{\to}V_j$ with $U,V\in\{P,R\}$, failure
pairs $U_i{\to}N_j$, recovery pairs $N_i{\to}U_j$, and within-failure pairs
$N_i{\to}N_j$. A same-index pair $P_i{\leftrightarrow}R_i$ has zero Hop and
explicitly supervises invariance, whereas cross-index robust pairs such as
$R_i{\to}R_j$ retain nonzero progress or regression according to $g_j-g_i$.

We train the GRM with mean-squared regression:
\begin{equation}
    \mathcal L_{\mathrm{pair}}(\theta)
    =
    \mathbb E_{(X_b,X_a)\sim\mathcal D}
    \left[
    (\hat y_{ba}-y_{ba})^2
    \right].
\end{equation}
We retain start- and goal-anchored comparisons and use fixed family quotas so
that abundant positive and within-branch pairs do not dominate robust, failure,
and recovery pairs.

\subsection{Signed-Hop Curriculum}
The signed Hop also provides a task-aligned measure of training difficulty. We
rank examples by $|\mathcal H_{ba}|$ within each pair family, so that frequent
families do not dominate selection. The large-Hop pool
$\mathcal D_{\mathrm{large}}$ contains the top 40\% of examples by
$|\mathcal H_{ba}|$ within each family, while
$\mathcal D_{\mathrm{fine}}$ contains the remaining small-Hop and zero-Hop
calibration examples.

Training uses two 200K-example pairwise-reward stages, totaling 400K pairwise
examples. Stage~1 samples from $\mathcal D_{\mathrm{large}}$ to learn global
progress, failure, and recovery geometry. Stage~2 primarily samples from
$\mathcal D_{\mathrm{fine}}$ to improve local calibration and robust invariance
while replaying a fraction $\rho$ of large-Hop examples:
\begin{equation}
\begin{aligned}
    \mathcal D_2
        ={}&
        (1-\rho)\mathcal D_{\mathrm{fine}}\\
        &+
        \rho\mathcal D_{\mathrm{large}}.
\end{aligned}
\label{eq:signed_hop_curriculum}
\end{equation}
We use $\rho=0.25$, so 25\% of Stage~2 examples are replayed large-Hop
examples and 75\% are fine or zero-Hop examples. The replay ratio changes only
the Stage~2 mixture; it does not alter the Stage~1 composition or the 40\%
large-Hop selection rule.

\subsection{Closed-Loop Reward Framework}
\label{sec:method_temporal_reward}
\label{sec:method_reward_inference}

At rollout step $t$, let $X_0$, $X_G=r_g$, $X_{t-1}$, and $X_t$ denote the
initial state, goal reference, previous observed state, and current observed
state. Online inference constructs temporal panels directly from observations
available in the current rollout and therefore requires neither a successful
source episode nor an endpoint-to-source alignment. To avoid repeatedly writing
the full GRM input, define the normalized query
\begin{equation}
\begin{aligned}
h_\theta&(X_b,X_a;\mathcal C)\\
&=
\operatorname{clip}\!\left(
\frac{
f_\theta(c,\mathcal R,\mathcal C,X_b,X_a)
}{100},
-1,1
\right).
\end{aligned}
\end{equation}

The online context is constructed separately for each query from the observed
rollout prefix. For the forward-anchored query $X_0{\to}X_t$, the pre-query
panel is empty and observations strictly between $X_0$ and $X_t$ populate the
between-state panels, yielding $\mathcal C_t^F$. For the incremental query
$X_{t-1}{\to}X_t$, observations before $X_{t-1}$ populate the pre-query panels,
while any observations between $X_{t-1}$ and $X_t$ populate the between-state
panels, yielding $\mathcal C_t^I$. The goal observation $X_G$ is a non-temporal task anchor, so the
backward-anchored query is evaluated without temporal panels. When an
explicitly aligned expert episode is available, the same slots may optionally
use its source-reference frames.

The three normalized predictions are therefore
\begin{equation}
\begin{aligned}
    \hat h_t^F
        &= h_\theta(X_0,X_t;\mathcal C_t^F),\\
    \hat h_t^B
        &= h_\theta(X_G,X_t;\varnothing),\\
    \hat h_t^I
        &= h_\theta(X_{t-1},X_t;\mathcal C_t^I).
\end{aligned}
\end{equation}
All online temporal panels contain only observations available before or at
rollout step $t$; no query uses future rollout observations. We denote the
resulting augmented online state by
$\bar s_t=(X_t,\mathcal C_t,m_t)$, where $\mathcal C_t$ collects the temporal
query context and $m_t$ denotes the recursive estimator memory.

Because the initial state has potential zero, a negative forward-anchored score
operationally indicates the negative branch. For $t\geq1$, we define
\begin{equation}
\begin{aligned}
    \hat b_t
        &= \mathbb I[\hat h_t^F<0],
    & \hat b_0
        &= 0,\\
    L_t^F
        &= L_t^B=-\hat b_t,
    & L_t^I
        &= -\max\{\hat b_{t-1},\hat b_t\}.
\end{aligned}
\end{equation}
Here $\hat b_t$ is the online branch gate: $L_t^F$ and $L_t^B$ depend on the
current branch, while $L_t^I$ depends on both adjacent branches so that a
negative-to-valid recovery pair retains the floor $-1$.

We invert a normalized Hop using
\begin{equation}
\Gamma(h;\phi,L)
=
\begin{cases}
\phi+h(1-\phi),
    & h\geq 0,\\
\phi+h(\phi-L),
    & h<0.
\end{cases}
\end{equation}
The three recovered potential estimates are
\begin{equation}
\begin{aligned}
    \hat\Phi_t^F
        &= \Gamma(\hat h_t^F;0,L_t^F),\\
    \hat\Phi_t^B
        &= \Gamma(\hat h_t^B;1,L_t^B),\\
    \hat\Phi_t^I
        &= \Gamma(
        \hat h_t^I;
        \hat\Phi_{t-1},
        L_t^I),
\end{aligned}
\end{equation}
and are fused as
\begin{equation}
    \hat\Phi_t
    =
    \operatorname{clip}\!\left(
    \frac{
    \hat\Phi_t^F+
    \hat\Phi_t^B+
    \hat\Phi_t^I
    }{3},
    -1,1
    \right),
    \qquad
    \hat\Phi_0=0.
\end{equation}

After inverse normalization, the three terms provide complementary estimates
of the context-conditioned online potential, denoted
$\hat\Phi_t\equiv\hat\Phi(\bar s_t)$. We retain the fixed unweighted
multi-anchor aggregation used by \citet{RoboDopamine2025}; the fusion weights
are neither learned nor selected on the evaluation set.

Finally, the reward supplied to the downstream RL algorithm is
\begin{equation}
    r_t^{\mathrm{GRM}}
    =
    r_t^{\mathrm{task}}
    +
    \gamma_{\mathrm{RL}}\hat\Phi_{t+1}
    -
    \hat\Phi_t,
    \label{eq:grm_shaping}
\end{equation}
where $r_t^{\mathrm{task}}$ is the original environment reward,
$\gamma_{\mathrm{RL}}$ is the discount factor used by policy optimization, and
$\hat\Phi_t=\hat\Phi(\bar s_t)$. With the GRM fixed during policy optimization,
using the same discount and assigning zero potential to the absorbing terminal
state retains the standard potential-based shaping form on the augmented
process~\citep{NgHaradaRussell1999}. The resulting potential is designed to be
insensitive to semantics-preserving appearance changes, responsive to
task-invalid failures, and positive for valid recovery transitions.

\section{Experiments and Results}
\label{sec:experiments}

\subsection{Experimental Setup}
\label{sec:training_evaluation}

\noindent\textbf{Training matrix.}
GRM-8B-Pro, released by \citet{RoboDopamine2025}, serves as a prior baseline;
Robo-Dopamine 2.0-8B uses Qwen3-VL-8B. We vary OOD supervision, disabled or
$2{\times}2$/$3{\times}3$ temporal context panels, prebuilt or runtime construction, and
visual-QA mixing. The main model uses 400K signed pairs with OOD supervision and prebuilt
$3{\times}3$ panels: standard pairs use same-episode expert history, whereas
synthetic OOD pairs use source-aligned successful-reference history. The RM is
not trained on downstream rollout trajectories. A fixed 200K visual-QA mix
(QA200K) is excluded from the signed-pair budget. Runtime variants construct
the same ordered panels at data-loading time and pad missing views. QA200K is
included identically across runs to preserve broad visual-language
understanding and instruction following.

\noindent\textbf{Curriculum controls.}
All Signed-Hop variants fix the Qwen3-VL-8B backbone~\citep{Qwen3VL2025}, OOD
supervision, prebuilt $3{\times}3$ panels with one between-state panel, token
budget, QA200K mix, and 400K signed pairs split evenly across two stages. We
compare random mixing (E0), the matched E4 pool without ordering (E1), reverse
fine-to-large training (E2), large-to-fine training without replay (E3), and
large-to-fine training with $\rho=0.25$ replay (E4). E1 isolates ordering; E3
isolates replay. The same QA200K auxiliary mix is fixed across curriculum
variants and excluded from the signed-pair budget; each curriculum stage
contains 200K examples.

\noindent\textbf{Evaluation protocol.}
We report trajectory-level visual order consistency (VOC): pairwise progress
predictions are accumulated along each trajectory and correlated with the
ground-truth order using Spearman correlation. Five families cover
in-distribution success (ID), temporal memory, OOD-negative, OOD-robust, and
OOD-temporal cases, evaluated with \texttt{static8} and
\texttt{history\_panel} inputs. Temporal cases include multi-stage,
multi-object, and mid-episode trajectories; OOD-temporal cases add edits such
as non-release. Temporal-memory examples come from dedicated temporal manifests
and trajectories explicitly marked as history-dependent. We report aggregate
VOC for each family rather than a per-cause breakdown. Together, these families
probe event-, phase-, and order-dependent progress instead of endpoint success
alone. Offline synthetic-OOD evaluation uses source-reference panels aligned by
dataset metadata, whereas simulation and real-robot RL use panels constructed
from the observed rollout history.

\subsection{Offline Results}
\label{sec:results}

\begin{table}[t]
\centering
{\small
\setlength{\tabcolsep}{3pt}
\begin{tabular}{@{}llcccccc@{}}
\toprule
Model & Input & ID & Temp. & Neg. & Rob. & OOD-T & Avg. \\
\midrule
GRM & static
& 0.974 & 0.977 & 0.951 & 0.939 & 0.948 & 0.958 \\
\midrule
Robo-Dopamine 2.0 & static
& 0.981 & 0.984 & 0.985 & 0.906 & 0.979 & 0.967 \\
Robo-Dopamine 2.0 & panel
& \textbf{0.991} & \textbf{0.994} & \textbf{0.994}
& \textbf{0.958} & \textbf{0.991} & \textbf{0.986} \\
\bottomrule
\end{tabular}
}
\caption{Trajectory-level VOC across five benchmark families. GRM denotes the
GRM-8B-Pro baseline; Robo-Dopamine 2.0 is one QA200K-mixed Qwen3-VL-8B checkpoint evaluated
with static or prebuilt $3{\times}3$ panel inputs. ID, Temp., Neg., Rob., and
OOD-T denote in-distribution, temporal-memory, OOD-negative, OOD-robust, and
OOD-temporal evaluation. Avg. is the unweighted mean; higher is better.}
\label{tab:voc}
\end{table}

\noindent Panels raise Robo-Dopamine 2.0 average VOC from 0.967 to 0.986 and OOD-robust
VOC from 0.906 to 0.958; GRM-8B-Pro reaches 0.958 average static VOC. The
same-checkpoint comparison isolates the benefit of training-aligned temporal
context from model capacity and training data.

\begin{figure}[t]
\centering
\includegraphics[width=0.98\columnwidth]{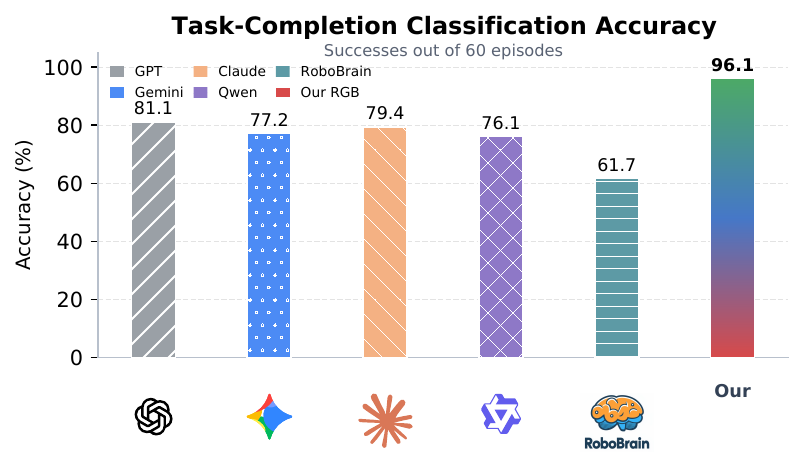}
\caption{Task-completion classification accuracy on 60 episodes. We compare
GPT-5.5, Gemini-3-Pro, Claude-Opus-4.8, Qwen3.5-397B-A17B-fp8, RoboBrain~2.0,
and Robo-Dopamine 2.0-8B. This discrete label metric is distinct from VOC.}
\label{fig:task_completion_accuracy}
\end{figure}

Robo-Dopamine 2.0-8B obtains 96.1\% task-completion accuracy, versus GPT-5.5 (81.1\%)
~\citep{OpenAI2026GPT55}, Claude-Opus-4.8 (79.4\%)
~\citep{Anthropic2026ClaudeOpus48}, Gemini-3-Pro (77.2\%)
~\citep{Google2025Gemini3Pro}, Qwen3.5-397B-A17B-fp8 (76.1\%)
~\citep{QwenTeam2026Qwen35}, and RoboBrain~2.0 (61.7\%)
~\citep{RoboBrain2025}.

\subsubsection{Temporal Context and OOD Ablations}
\label{sec:ablation}

\begin{table}[t]
\centering
{\small
\begin{tabular}{@{}llccc@{}}
\toprule
Base & Training condition & Static & Panel & Gain \\
\midrule
GRM & baseline & 0.958 & 0.965 & +0.007 \\
Robo-Dopamine 2.0 & $3{\times}3$+QA & 0.967 & 0.986 & +0.019 \\
\midrule
GRM & OOD+no panel+QA & 0.967 & 0.977 & +0.010 \\
GRM & OOD+$2{\times}2$ & 0.957 & 0.985 & +0.028 \\
GRM & OOD+$3{\times}3$ & 0.955 & \textbf{0.988} & +0.033 \\
Robo-Dopamine 2.0 & OOD+no panel & 0.965 & 0.974 & +0.009 \\
Robo-Dopamine 2.0 & OOD+$2{\times}2$ & 0.962 & 0.985 & +0.023 \\
Robo-Dopamine 2.0 & OOD+$3{\times}3$ & 0.960 & \textbf{0.987} & +0.027 \\
\bottomrule
\end{tabular}
}
\caption{Temporal-context and OOD ablations in average VOC. Static and Panel query the
same checkpoint; $+\mathrm{QA}$ denotes QA200K. Full results are supplementary.}
\label{tab:memory_ablation}
\end{table}

\noindent Structured panels provide the largest gain: moving from OOD no-panel
training to $3{\times}3$ panels raises panel VOC from 0.977 to 0.988 for GRM and
from 0.974 to 0.987 for Robo-Dopamine 2.0. OOD supervision also improves panel VOC
over no-OOD, no-panel training, but the panel contribution is larger. QA200K
acts as a weak regularizer for no-panel models, with a smaller marginal effect
once structured temporal context is available. Across
backbones, one or two between-state panels perform best; larger panels are not
monotonic under a fixed visual-token budget, and prebuilt panels outperform
runtime construction. The non-monotonic trend reflects a trade-off between
additional temporal evidence and the visual-token budget. Runtime construction
also changes panel provenance, image order, and prompt tokens relative to the
training distribution. Full panel-size, panel-count, and construction
factorials are provided in the supplementary material.

\subsubsection{Signed-Hop Curriculum}
\label{sec:curriculum_results}

Signed-Hop first learns global geometry from large-Hop pairs, then refines local
calibration with small and zero-Hop pairs while replaying large-Hop examples.
Under the fixed 400K budget, 25\% replay reaches 0.9872 average VOC, exceeding
the matched-pool shuffled control (0.9858) and no replay (0.9866). The shuffled
control uses exactly the same sample pool as the replay curriculum, isolating
ordering from data composition; the no-replay variant isolates retention of
large-Hop geometry during local refinement.

\begin{figure}[t]
\centering
\includegraphics[width=0.75\columnwidth]{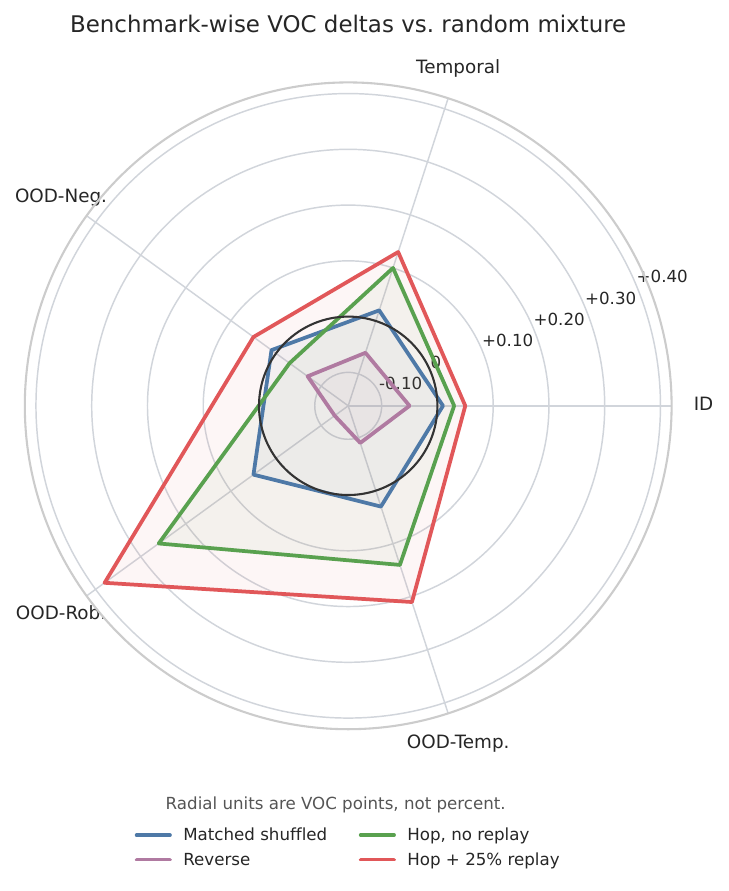}
\caption{Per-family VOC gains of Signed-Hop over its matched-pool shuffled
control. Replay sensitivity is provided in the supplementary material.}
\label{fig:signed_hop_ablation}
\end{figure}

Figure~\ref{fig:signed_hop_ablation} shows gains in every family, especially
OOD-robust and OOD-temporal cases. Curriculum gains complement the larger gains
from temporal context panels. Signed-Hop is training-only: large-Hop pairs establish
global progress, failure, and recovery geometry before small and zero-Hop pairs
refine local calibration. The full replay sweep and branch-wise diagnostics are
provided in the supplementary material.

\subsection{Reinforcement Learning}
\label{sec:rl_experiments}

In all downstream RL experiments, the RM remains frozen and the history-enabled
conditions populate the same panel slots causally from the observed rollout
prefix; no source alignment or repetition labels are used.

\subsubsection{Simulation on RoboTwin}
\label{sec:simulation_rl_results}

We optimize task-specific OpenVLA-OFT~\citep{kim2025oft} policies with
GRPO~\citep{Shao2024DeepSeekMath} in RLinf~\citep{zang2025rlinf} on three
RoboTwin~2.0 tasks~\citep{chen2025robotwin}: \texttt{place\_empty\_cup},
\texttt{place\_container\_plate}, and \texttt{handover\_block}. Conditions
share policy initialization, environment, optimizer, and evaluation protocol,
so the controlled factorial tests whether OOD supervision and temporal context
improve policy learning. RLinf Sparse uses only a termination-triggered success
reward (scale 5) for 1,000 epochs; C00--C11 use the shaping interface in
Eq.~\eqref{eq:grm_shaping} for 300 epochs. C00 enables neither OOD supervision
nor rollout-history panels, C01 adds rollout-history panels, C10 adds OOD
supervision, and C11 uses both. Complete policy, GRPO, evaluation,
and task-specific hyperparameters are supplementary.

\begin{table}[t]
\centering
{\small
\setlength{\tabcolsep}{3pt}
\begin{tabular}{@{}lcccccc@{}}
\toprule
Reward condition & OOD & Hist. & Cup & Plate & Hand. & Avg. \\
\midrule
RLinf Sparse & -- & -- & 94.2 & \textbf{95.0} & 68.2 & 85.8 \\
C00 Controlled & $\times$ & $\times$ & 77.5 & 82.1 & 57.0 & 72.2 \\
C01 History & $\times$ & $\checkmark$ & 79.5 & 88.6 & 57.1 & 75.1 \\
C10 OOD & $\checkmark$ & $\times$ & 85.2 & 89.5 & 67.1 & 80.6 \\
C11 Full & $\checkmark$ & $\checkmark$ & \textbf{96.2} & 94.6 & \textbf{69.7} & \textbf{86.8} \\
\bottomrule
\end{tabular}
}
\caption{RoboTwin success rate (\%). Sparse uses 1,000 epochs; C00--C11 use
300. Avg. is the unweighted task mean.}
\label{tab:robotwin_rl}
\end{table}

Within 300 epochs, rollout-history panels raise mean success from 72.2\% (C00) to
75.1\% (C01), OOD supervision raises it to 80.6\% (C10), and their combination
reaches 86.8\% (C11). C11 exceeds the 1,000-epoch sparse baseline (85.8\%)
overall and on two of three tasks, supporting complementary gains from
rollout history and OOD supervision. Relative to C10 and C01, C11 improves
the mean by 6.2 and 11.7 points, respectively. It is best on
\texttt{place\_empty\_cup} and \texttt{handover\_block}; the sparse baseline
remains 0.4 points higher on \texttt{place\_container\_plate}.

\subsubsection{Real-World Repeated Insertions}
\label{sec:real_robot}

We evaluate ConRFT~\citep{chen2025conrft} on dual Franka arms with multi-view
RGB observations. In \emph{Insert Square}, one arm aligns a four-hole block
with four pegs for $K=4$ consecutive attempts. Event Sparse uses a reward after
each insertion; C00--C11 add GRM shaping. A terminal-only pilot did not learn.
C00--C11 use 200 online episodes and Event Sparse uses 300; all conditions use
20 evaluation trials and shared controls. Details are supplementary.

Repeated insertion produces visually similar endpoints at different cumulative
phases. The observed rollout history supplies event and phase information for
interpreting recurring endpoint observations. Matched real-robot traces show
non-monotonic progress jumps with static queries and more order-consistent
estimates with history panels; these
traces are provided in the supplementary material.

\begin{figure}[t]
    \centering
    \includegraphics[width=\columnwidth]{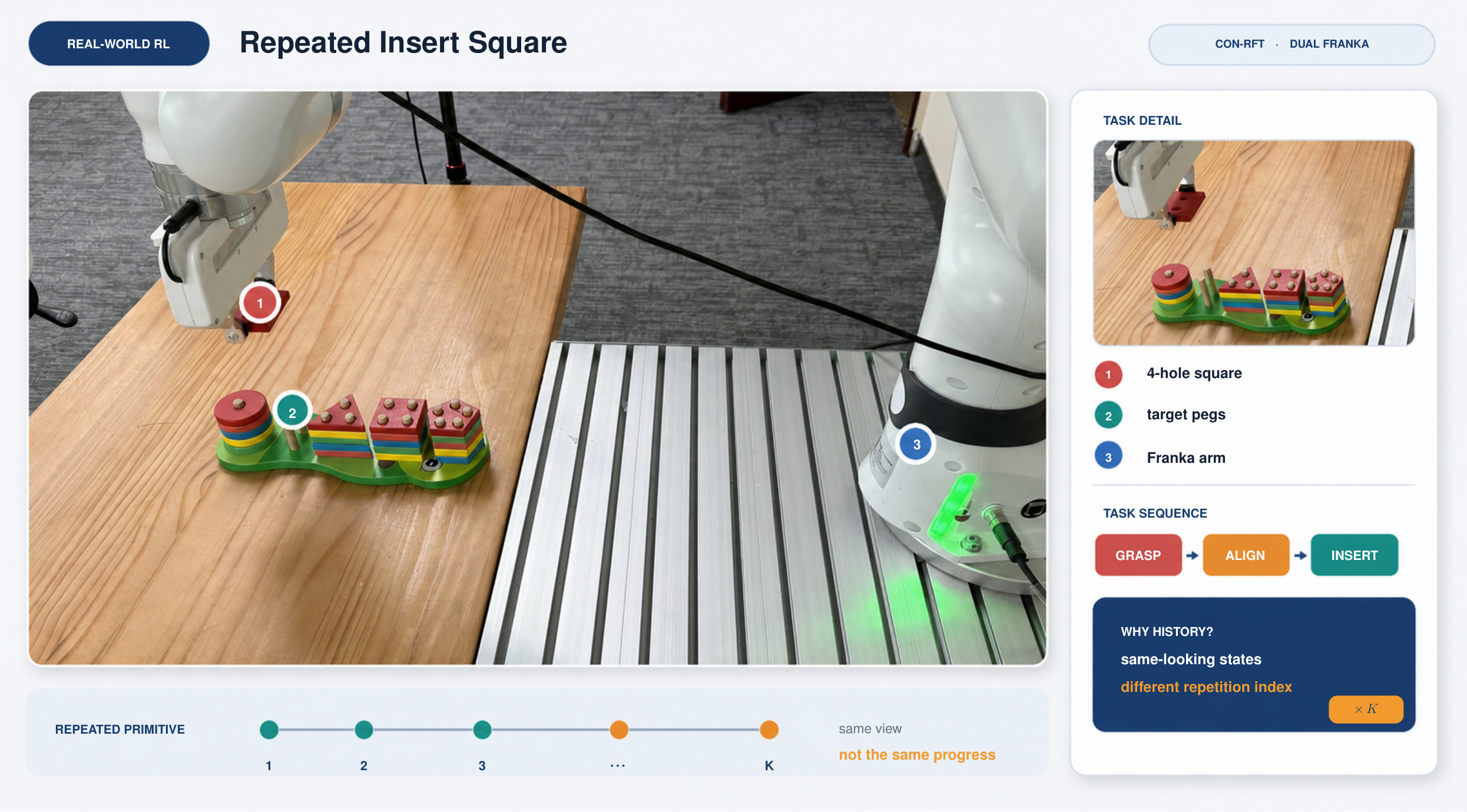}
    \caption{Dual-Franka repeated \emph{Insert Square} setup with ConRFT. The
    numbered callouts identify the four-hole square, target pegs, and acting arm.
    Each trial has $K=4$ attempts, creating visually similar states at different
    cumulative progress.}
    \label{fig:real_robot_insert_square}
\end{figure}

\begin{table}[t]
\centering
{\small
\begin{tabular}{@{}lcccc@{}}
\toprule
Reward & OOD & Hist. & Inserts / 80 & Full / 20 \\
\midrule
Event Sparse & -- & -- & 48 & 8 \\
C00 Controlled & $\times$ & $\times$ & 46 & 7 \\
C01 History & $\times$ & $\checkmark$ & 67 & 13 \\
C10 OOD & $\checkmark$ & $\times$ & 50 & 9 \\
C11 Full & $\checkmark$ & $\checkmark$ & \textbf{71} & \textbf{15} \\
\bottomrule
\end{tabular}
}
\caption{Repeated \emph{Insert Square}. Event Sparse uses only the per-insertion
event reward; C00--C11 additionally use GRM shaping. Inserts counts successes
over 80 attempts; Full counts 20 trials completing all four attempts.}
\label{tab:real_insert_square}
\end{table}

Rollout-history panels provide the largest gain: relative to C00 and C10, C01 and C11
each add 21 successful insertions and 6 full-sequence successes. C11 reaches 71/80
insertions and 15/20 full successes, versus 48/80 and 8/20 for Event Sparse,
using 200 rather than 300 online episodes.

\section{Conclusion}

\method{} combines history-conditioned pairwise prediction, an OOD-aware signed
progress space, and a Signed-Hop curriculum in one reward interface. Standard
queries use same-episode expert context, synthetic OOD queries retain edited
endpoints with aligned successful-reference panels, and online queries use
observed rollout history. The signed space models valid progress,
robustness-preserving variations, task-invalid failures, and recovery, while
the curriculum learns large transitions before fine-grained calibration.
Offline and downstream RL results show complementary gains from temporal
context, OOD supervision, curriculum ordering, and replay across offline
ordering, simulation, and real-world manipulation settings. The same pairwise
interface supports offline progress estimation and closed-loop reward shaping.
Future work will broaden task, OOD, and real-world evaluation.

\bibliography{references}

@inproceedings{RoboDopamine2025,
  author = {Tan, Huajie and Chen, Sixiang and Xu, Yijie and Wang, Zixiao and Chi, Cheng and Ji, Yuheng and Lyu, Yaoxu and Zhao, Zhongxia and Chen, Xiansheng and Co, Peterson and Xie, Shaoxuan and Yao, Guocai and Wang, Pengwei and Wang, Zhongyuan and Zhang, Shanghang},
  title = {General Process Reward Modeling for Robotic Reinforcement Learning},
  booktitle = {Proceedings of the IEEE/CVF Conference on Computer Vision and Pattern Recognition ({CVPR})},
  pages = {22412--22422},
  month = jun,
  year = {2026}
}

@misc{chen2025conrft,
  author = {Chen, Yuhui and Tian, Shuai and Liu, Shugao and Zhou, Yingting and Li, Haoran and Zhao, Dongbin},
  title = {{ConRFT}: A Reinforced Fine-Tuning Method for {VLA} Models via Consistency Policy},
  year = {2025},
  eprint = {2502.05450},
  archivePrefix = {arXiv},
  primaryClass = {cs.RO},
  doi = {10.48550/arXiv.2502.05450},
  url = {https://arxiv.org/abs/2502.05450}
}

@inproceedings{NgHaradaRussell1999,
  author = {Ng, Andrew Y. and Harada, Daishi and Russell, Stuart J.},
  title = {Policy Invariance Under Reward Transformations: Theory and Application to Reward Shaping},
  booktitle = {Proceedings of the Sixteenth International Conference on Machine Learning},
  pages = {278--287},
  publisher = {Morgan Kaufmann},
  address = {San Francisco, CA, USA},
  year = {1999}
}

@inproceedings{RT12022,
  author = {Brohan, Anthony and Brown, Noah and Carbajal, Justice and Chebotar, Yevgen and Dabis, Joseph and Finn, Chelsea and Gopalakrishnan, Keerthana and Hausman, Karol and Herzog, Alexander and Hsu, Jasmine and Ibarz, Julian and Ichter, Brian and Irpan, Alex and Jackson, Tomas and Jesmonth, Sally and Joshi, Nikhil and Julian, Ryan and Kalashnikov, Dmitry and Kuang, Yuheng and Leal, Isabel and Lee, Kuang-Huei and Levine, Sergey and Lu, Yao and Malla, Utsav and Manjunath, Deeksha and Mordatch, Igor and Nachum, Ofir and Parada, Carolina and Peralta, Jodilyn and Perez, Emily and Pertsch, Karl and Quiambao, Jornell and Rao, Kanishka and Ryoo, Michael S. and Salazar, Grecia and Sanketi, Pannag R. and Sayed, Kevin and Singh, Jaspiar and Sontakke, Sumedh and Stone, Austin and Tan, Clayton and Tran, Huong and Vanhoucke, Vincent and Vega, Steve and Vuong, Quan H. and Xia, Fei and Xiao, Ted and Xu, Peng and Xu, Sichun and Yu, Tianhe and Zitkovich, Brianna},
  title = {{RT-1}: Robotics Transformer for Real-World Control at Scale},
  booktitle = {Proceedings of Robotics: Science and Systems},
  address = {Daegu, Republic of Korea},
  month = jul,
  year = {2023},
  doi = {10.15607/RSS.2023.XIX.025}
}

@inproceedings{RT22023,
  author = {Zitkovich, Brianna and Yu, Tianhe and Xu, Sichun and Xu, Peng and Xiao, Ted and Xia, Fei and Wu, Jialin and Wohlhart, Paul and Welker, Stefan and Wahid, Ayzaan and Vuong, Quan and Vanhoucke, Vincent and Tran, Huong and Soricut, Radu and Singh, Anikait and Singh, Jaspiar and Sermanet, Pierre and Sanketi, Pannag R. and Salazar, Grecia and Ryoo, Michael S. and Reymann, Krista and Rao, Kanishka and Pertsch, Karl and Mordatch, Igor and Michalewski, Henryk and Lu, Yao and Levine, Sergey and Lee, Lisa and Lee, Tsang-Wei Edward and Leal, Isabel and Kuang, Yuheng and Kalashnikov, Dmitry and Julian, Ryan and Joshi, Nikhil J. and Irpan, Alex and Ichter, Brian and Hsu, Jasmine and Herzog, Alexander and Hausman, Karol and Gopalakrishnan, Keerthana and Fu, Chuyuan and Florence, Pete and Finn, Chelsea and Dubey, Kumar Avinava and Driess, Danny and Ding, Tianli and Choromanski, Krzysztof Marcin and Chen, Xi and Chebotar, Yevgen and Carbajal, Justice and Brown, Noah and Brohan, Anthony and Arenas, Montserrat Gonzalez and Han, Kehang},
  title = {{RT-2}: Vision-Language-Action Models Transfer Web Knowledge to Robotic Control},
  booktitle = {Proceedings of the 7th Conference on Robot Learning},
  pages = {2165--2183},
  editor = {Tan, Jie and Toussaint, Marc and Darvish, Kourosh},
  volume = {229},
  series = {Proceedings of Machine Learning Research},
  publisher = {PMLR},
  year = {2023}
}

@inproceedings{OpenVLA2024,
  author = {Kim, Moo Jin and Pertsch, Karl and Karamcheti, Siddharth and Xiao, Ted and Balakrishna, Ashwin and Nair, Suraj and Rafailov, Rafael and Foster, Ethan P. and Sanketi, Pannag R. and Vuong, Quan and Kollar, Thomas and Burchfiel, Benjamin and Tedrake, Russ and Sadigh, Dorsa and Levine, Sergey and Liang, Percy and Finn, Chelsea},
  title = {{OpenVLA}: An Open-Source Vision-Language-Action Model},
  booktitle = {Proceedings of the 8th Conference on Robot Learning},
  pages = {2679--2713},
  editor = {Agrawal, Pulkit and Kroemer, Oliver and Burgard, Wolfram},
  volume = {270},
  series = {Proceedings of Machine Learning Research},
  publisher = {PMLR},
  year = {2025}
}

@inproceedings{kim2025oft,
  author = {Kim, Moo Jin and Finn, Chelsea and Liang, Percy},
  title = {Fine-Tuning Vision-Language-Action Models: Optimizing Speed and Success},
  booktitle = {Proceedings of Robotics: Science and Systems},
  address = {Los Angeles, CA, USA},
  month = jun,
  year = {2025},
  doi = {10.15607/RSS.2025.XXI.017}
}

@misc{zang2025rlinf,
  author = {Yu, Chao and Wang, Yuanqing and Guo, Zhen and Lin, Hao and Xu, Si and Zang, Hongzhi and others},
  title = {{RLinf}: Flexible and Efficient Large-Scale Reinforcement Learning via Macro-to-Micro Flow Transformation},
  year = {2025},
  eprint = {2509.15965},
  archivePrefix = {arXiv},
  primaryClass = {cs.DC},
  doi = {10.48550/arXiv.2509.15965},
  url = {https://arxiv.org/abs/2509.15965}
}

@inproceedings{chen2025robotwin,
  author = {Chen, Tianxing and Chen, Zanxin and Chen, Baijun and Cai, Zijian and Liu, Yibin and Li, Zixuan and Liang, Qiwei and Lin, Xianliang and Ge, Yiheng and Gu, Zhenyu and Deng, Weiliang and Guo, Yubin and Nian, Tian and Xie, Xuanbing and Chen, Qiangyu and Su, Kailun and Xu, Tianling and Liu, Guodong and Hu, Mengkang and Gao, Huan-ang and Wang, Kaixuan and Liang, Zhixuan and Qin, Yusen and Yang, Xiaokang and Luo, Ping and Mu, Yao},
  title = {{RoboTwin} 2.0: A Scalable Data Generator and Benchmark with Strong Domain Randomization for Robust Bimanual Robotic Manipulation},
  booktitle = {Proceedings of the 43rd International Conference on Machine Learning},
  year = {2026}
}

@inproceedings{RoboCasa2024,
  author = {Nasiriany, Soroush and Maddukuri, Abhiram and Zhang, Lance and Parikh, Adeet and Lo, Aaron and Joshi, Abhishek and Mandlekar, Ajay and Zhu, Yuke},
  title = {{RoboCasa}: Large-Scale Simulation of Household Tasks for Generalist Robots},
  booktitle = {Proceedings of Robotics: Science and Systems},
  year = {2024},
  doi = {10.15607/RSS.2024.XX.050}
}

@inproceedings{LIBERO2023,
  author = {Liu, Bo and Zhu, Yifeng and Gao, Chongkai and Feng, Yihao and Liu, Qiang and Zhu, Yuke and Stone, Peter},
  title = {{LIBERO}: Benchmarking Knowledge Transfer for Lifelong Robot Learning},
  booktitle = {Advances in Neural Information Processing Systems},
  volume = {36},
  year = {2023}
}

@misc{AgiBotWorld2025,
  author = {{AgiBot-World Contributors} and others},
  title = {AgiBot World Colosseo: A Large-Scale Manipulation Platform for Scalable and Intelligent Embodied Systems},
  year = {2025},
  eprint = {2503.06669},
  archivePrefix = {arXiv},
  primaryClass = {cs.RO},
  doi = {10.48550/arXiv.2503.06669}
}

@inproceedings{LIV2023,
  author = {Ma, Yecheng Jason and Kumar, Vikash and Zhang, Amy and Bastani, Osbert and Jayaraman, Dinesh},
  title = {{LIV}: Language-Image Representations and Rewards for Robotic Control},
  booktitle = {Proceedings of the 40th International Conference on Machine Learning},
  pages = {23301--23320},
  editor = {Krause, Andreas and Brunskill, Emma and Cho, Kyunghyun and Engelhardt, Barbara and Sabato, Sivan and Scarlett, Jonathan},
  volume = {202},
  series = {Proceedings of Machine Learning Research},
  publisher = {PMLR},
  year = {2023}
}

@misc{VLMRewards2023,
  author = {Baumli, Kate and Baveja, Satinder and Behbahani, Feryal and Chan, Harris and Comanici, Gheorghe and Flennerhag, Sebastian and Gazeau, Maxime and Holsheimer, Kristian and Horgan, Dan and Laskin, Michael and Lyle, Clare and Masoom, Hussain and McKinney, Kay and Mnih, Volodymyr and Neitz, Alexander and Nikulin, Dmitry and Pardo, Fabio and Parker-Holder, Jack and Quan, John and Rockt{\"a}schel, Tim and Sahni, Himanshu and Schaul, Tom and Schroecker, Yannick and Spencer, Stephen and Steigerwald, Richie and Wang, Luyu and Zhang, Lei},
  title = {Vision-Language Models as a Source of Rewards},
  year = {2023},
  eprint = {2312.09187},
  archivePrefix = {arXiv}
}

@inproceedings{VICtoR2024,
  author = {Hung, Kuo-Han and Lo, Pang-Chi and Yeh, Jia-Fong and Hsu, Han-Yuan and Chen, Yi-Ting and Hsu, Winston H.},
  title = {{VICtoR}: Learning Hierarchical Vision-Instruction Correlation Rewards for Long-horizon Manipulation},
  booktitle = {The Thirteenth International Conference on Learning Representations},
  year = {2025}
}

@misc{RoboReward2026,
  author = {Lee, Tony and Wagenmaker, Andrew and Pertsch, Karl and Liang, Percy and Levine, Sergey and Finn, Chelsea},
  title = {{RoboReward}: General-Purpose Vision-Language Reward Models for Robotics},
  year = {2026},
  eprint = {2601.00675},
  archivePrefix = {arXiv}
}

@inproceedings{VLAC2025,
  author = {Zhang, Qi and Zhai, Shaopeng and Zhang, Shengzhe and Liu, Litao and Zhang, Tianyi and Huang, Fuxian and Zhou, Ming},
  title = {A Generalist Pair-wise Progress Critic Model for Vision-Language-Action Robots},
  booktitle = {Proceedings of the 43rd International Conference on Machine Learning},
  year = {2026}
}

@inproceedings{SARM2025,
  author = {Chen, Qianzhong and Yu, Justin and Schwager, Mac and Abbeel, Pieter and Shentu, Fred and Wu, Philipp},
  title = {{SARM}: Stage-Aware Reward Modeling for Long Horizon Robot Manipulation},
  booktitle = {The Fourteenth International Conference on Learning Representations},
  year = {2026}
}

@inproceedings{ARM2026,
  author = {Mao, Yiming and Yu, Zixi and Mao, Weixin and Li, Yinhao and Hu, Qirui and Lan, Zihan and Zhu, Minzhao and Chen, Hua},
  title = {{ARM}: Advantage Reward Modeling for Long-Horizon Manipulation},
  booktitle = {Proceedings of the IEEE/CVF Conference on Computer Vision and Pattern Recognition ({CVPR}) Workshops},
  pages = {4468--4477},
  month = jun,
  year = {2026}
}

@inproceedings{AHA2024,
  author = {Duan, Jiafei and Pumacay, Wilbert and Kumar, Nishanth and Wang, Yi Ru and Tian, Shulin and Yuan, Wentao and Krishna, Ranjay and Fox, Dieter and Mandlekar, Ajay and Guo, Yijie},
  title = {{AHA}: A Vision-Language-Model for Detecting and Reasoning Over Failures in Robotic Manipulation},
  booktitle = {The Thirteenth International Conference on Learning Representations},
  year = {2025}
}

@misc{MARVL2026,
  author = {Zhou, Xunlan and Chen, Xuanlin and Zhang, Shaowei and Li, Xiangkun and Wan, ShengHua and Hu, Xiaohai and Yuan, Lei and Gan, Le and Zhan, De-chuan},
  title = {{MARVL}: Multi-Stage Guidance for Robotic Manipulation via Vision-Language Models},
  year = {2026},
  eprint = {2602.15872},
  archivePrefix = {arXiv}
}

@inproceedings{PEBBLE2021,
  author = {Lee, Kimin and Smith, Laura M. and Abbeel, Pieter},
  title = {{PEBBLE}: Feedback-Efficient Interactive Reinforcement Learning via Relabeling Experience and Unsupervised Pre-Training},
  booktitle = {Proceedings of the 38th International Conference on Machine Learning},
  pages = {6152--6163},
  editor = {Meila, Marina and Zhang, Tong},
  volume = {139},
  series = {Proceedings of Machine Learning Research},
  publisher = {PMLR},
  year = {2021},
  url = {https://proceedings.mlr.press/v139/lee21i.html}
}

@inproceedings{RLVLMF2024,
  author = {Wang, Yufei and Sun, Zhanyi and Zhang, Jesse and Xian, Zhou and Biyik, Erdem and Held, David and Erickson, Zackory},
  title = {{RL-VLM-F}: Reinforcement Learning from Vision Language Foundation Model Feedback},
  booktitle = {Proceedings of the 41st International Conference on Machine Learning},
  pages = {51484--51501},
  editor = {Salakhutdinov, Ruslan and Kolter, Zico and Heller, Katherine and Weller, Adrian and Oliver, Nuria and Scarlett, Jonathan and Berkenkamp, Felix},
  volume = {235},
  series = {Proceedings of Machine Learning Research},
  publisher = {PMLR},
  year = {2024},
  url = {https://proceedings.mlr.press/v235/wang24bn.html}
}

@inproceedings{RoboCLIP2023,
  author = {Sontakke, Sumedh and Zhang, Jesse and Arnold, S{\'e}b and Pertsch, Karl and B{\i}y{\i}k, Erdem and Sadigh, Dorsa and Finn, Chelsea and Itti, Laurent},
  title = {{RoboCLIP}: One Demonstration is Enough to Learn Robot Policies},
  booktitle = {Advances in Neural Information Processing Systems},
  volume = {36},
  pages = {55681--55693},
  year = {2023},
  doi = {10.52202/075280-2430}
}

@inproceedings{VIP2023,
  author = {Ma, Yecheng Jason and Sodhani, Shagun and Jayaraman, Dinesh and Bastani, Osbert and Kumar, Vikash and Zhang, Amy},
  title = {{VIP}: Towards Universal Visual Reward and Representation via Value-Implicit Pre-Training},
  booktitle = {The Eleventh International Conference on Learning Representations},
  year = {2023},
  url = {https://openreview.net/forum?id=YJ7o2wetJ2}
}

@inproceedings{LOReL2022,
  author = {Nair, Suraj and Mitchell, Eric and Chen, Kevin and Ichter, Brian and Savarese, Silvio and Finn, Chelsea},
  title = {Learning Language-Conditioned Robot Behavior from Offline Data and Crowd-Sourced Annotation},
  booktitle = {Proceedings of the 5th Conference on Robot Learning},
  pages = {1303--1315},
  editor = {Faust, Aleksandra and Hsu, David and Neumann, Gerhard},
  volume = {164},
  series = {Proceedings of Machine Learning Research},
  publisher = {PMLR},
  year = {2022},
  url = {https://proceedings.mlr.press/v164/nair22a.html}
}

@inproceedings{Ni2022RecurrentPOMDP,
  author = {Ni, Tianwei and Eysenbach, Benjamin and Salakhutdinov, Ruslan},
  title = {Recurrent Model-Free {RL} Can Be a Strong Baseline for Many {POMDP}s},
  booktitle = {Proceedings of the 39th International Conference on Machine Learning},
  pages = {16691--16723},
  editor = {Chaudhuri, Kamalika and Jegelka, Stefanie and Song, Le and Szepesvari, Csaba and Niu, Gang and Sabato, Sivan},
  volume = {162},
  series = {Proceedings of Machine Learning Research},
  publisher = {PMLR},
  year = {2022},
  url = {https://proceedings.mlr.press/v162/ni22a.html}
}

@inproceedings{Guhur2023HistoryAware,
  author = {Guhur, Pierre-Louis and Chen, Shizhe and Pinel, Ricardo Garcia and Tapaswi, Makarand and Laptev, Ivan and Schmid, Cordelia},
  title = {Instruction-Driven History-Aware Policies for Robotic Manipulations},
  booktitle = {Proceedings of the 6th Conference on Robot Learning},
  pages = {175--187},
  editor = {Liu, Karen and Kulic, Dana and Ichnowski, Jeff},
  volume = {205},
  series = {Proceedings of Machine Learning Research},
  publisher = {PMLR},
  year = {2023},
  url = {https://proceedings.mlr.press/v205/guhur23a.html}
}

@inproceedings{Fu2024TemporalOT,
  author = {Fu, Yuwei and Zhang, Haichao and Wu, Di and Xu, Wei and Boulet, Benoit},
  title = {Robot Policy Learning with Temporal Optimal Transport Reward},
  booktitle = {Advances in Neural Information Processing Systems},
  volume = {37},
  pages = {122078--122103},
  year = {2024},
  doi = {10.52202/079017-3879}
}

@inproceedings{Adapt2Reward2024,
  author = {Yang, Yanting and Chen, Minghao and Qiu, Qibo and Wu, Jiahao and Wang, Wenxiao and Lin, Binbin and Guan, Ziyu and He, Xiaofei},
  title = {{Adapt2Reward}: Adapting Video-Language Models to Generalizable Robotic Rewards via Failure Prompts},
  booktitle = {Computer Vision -- {ECCV} 2024},
  pages = {163--180},
  publisher = {Springer Nature Switzerland},
  year = {2024},
  doi = {10.1007/978-3-031-72998-0_10}
}

@misc{VideoLanguageCritic2024,
  author = {Alakuijala, Minttu and McLean, Reginald and Woungang, Isaac and Farsad, Nariman and Kaski, Samuel and Marttinen, Pekka and Yuan, Kai},
  title = {Video-Language Critic: Transferable Reward Functions for Language-Conditioned Robotics},
  year = {2024},
  eprint = {2405.19988},
  archivePrefix = {arXiv},
  primaryClass = {cs.RO},
  doi = {10.48550/arXiv.2405.19988},
  url = {https://arxiv.org/abs/2405.19988}
}

@misc{Qwen3VL2025,
  author = {Bai, Shuai and Cai, Yuxuan and Chen, Ruizhe and Chen, Keqin and Chen, Xionghui and Cheng, Zesen and Deng, Lianghao and Ding, Wei and Gao, Chang and Ge, Chunjiang and Ge, Wenbin and Guo, Zhifang and Huang, Qidong and Huang, Jie and Huang, Fei and Hui, Binyuan and Jiang, Shutong and Li, Zhaohai and Li, Mingsheng and Li, Mei and Li, Kaixin and Lin, Zicheng and Lin, Junyang and Liu, Xuejing and Liu, Jiawei and Liu, Chenglong and Liu, Yang and Liu, Dayiheng and Liu, Shixuan and Lu, Dunjie and Luo, Ruilin and Lv, Chenxu and Men, Rui and Meng, Lingchen and Ren, Xuancheng and Ren, Xingzhang and Song, Sibo and Sun, Yuchong and Tang, Jun and Tu, Jianhong and Wan, Jianqiang and Wang, Peng and Wang, Pengfei and Wang, Qiuyue and Wang, Yuxuan and Xie, Tianbao and Xu, Yiheng and Xu, Haiyang and Xu, Jin and Yang, Zhibo and Yang, Mingkun and Yang, Jianxin and Yang, An and Yu, Bowen and Zhang, Fei and Zhang, Hang and Zhang, Xi and Zheng, Bo and Zhong, Humen and Zhou, Jingren and Zhou, Fan and Zhou, Jing and Zhu, Yuanzhi and Zhu, Ke},
  title = {{Qwen3-VL} Technical Report},
  year = {2025},
  eprint = {2511.21631},
  archivePrefix = {arXiv}
}

@misc{OpenAI2026GPT55,
  author = {{OpenAI}},
  title = {{GPT-5.5} System Card},
  year = {2026},
  howpublished = {\url{https://openai.com/index/gpt-5-5-system-card/}},
  note = {Accessed: 2026-07-24}
}

@misc{Google2025Gemini3Pro,
  author = {{Google}},
  title = {{Gemini 3 Pro Preview}},
  year = {2025},
  howpublished = {\url{https://ai.google.dev/gemini-api/docs/models/gemini-3-pro-preview}},
  note = {Model ID: gemini-3-pro-preview. Accessed: 2026-07-24}
}

@misc{Anthropic2026ClaudeOpus48,
  author = {{Anthropic}},
  title = {Introducing {Claude Opus 4.8}},
  year = {2026},
  howpublished = {\url{https://www.anthropic.com/research/claude-opus-4-8}},
  note = {Model ID: claude-opus-4-8. Accessed: 2026-07-24}
}

@misc{QwenTeam2026Qwen35,
  author = {{Qwen Team}},
  title = {{Qwen3.5-397B-A17B-FP8}},
  year = {2026},
  howpublished = {\url{https://huggingface.co/Qwen/Qwen3.5-397B-A17B-FP8}},
  note = {Accessed: 2026-07-24}
}

@misc{RoboBrain2025,
  author = {{BAAI RoboBrain Team} and Cao, Mingyu and Tan, Huajie and Ji, Yuheng and Lin, Minglan and Li, Zhiyu and Cao, Zhou and Wang, Pengwei and Zhou, Enshen and Han, Yi and Tang, Yingbo and Xu, Xiangqi and Guo, Wei and Lyu, Yaoxu and Xu, Yijie and Shi, Jiayu and Du, Mengfei and Chi, Cheng and Zhao, Mengdi and Hao, Xiaoshuai and Zhao, Junkai and Zhang, Xiaojie and Rong, Shanyu and Lyu, Huaihai and Cai, Zhengliang and Fu, Yankai and Chen, Ning and Zhang, Bolun and Zhang, Lingfeng and Zhang, Shuyi and Liu, Dong and Feng, Xi and Wang, Songjing and Liu, Xiaodan and Jiao, Yance and Lyu, Mengsi and Chen, Zhuo and He, Chenrui and Ao, Yulong and Sun, Xue and He, Zheqi and Zheng, Jingshu and Yang, Xi and Shi, Donghai and Xie, Kunchang and Zhang, Bochao and Nie, Shaokai and Men, Chunlei and Lin, Yonghua and Wang, Zhongyuan and Huang, Tiejun and Zhang, Shanghang},
  title = {{RoboBrain} 2.0 Technical Report},
  year = {2025},
  eprint = {2507.02029},
  archivePrefix = {arXiv}
}

@misc{Shao2024DeepSeekMath,
  author = {Shao, Zhihong and Wang, Peiyi and Zhu, Qihao and Xu, Runxin and Song, Junxiao and Bi, Xiao and Zhang, Haowei and Zhang, Mingchuan and Li, Y. K. and Wu, Y. and Guo, Daya},
  title = {{DeepSeekMath}: Pushing the Limits of Mathematical Reasoning in Open Language Models},
  year = {2024},
  eprint = {2402.03300},
  archivePrefix = {arXiv}
}
\clearpage
\urlstyle{rm} 
\def\UrlFont{\rm}  
\frenchspacing  

\appendix
\section*{Supplementary Material}

\section{OOD Data Construction Pipeline}
\label{app:ood_pipeline}

We construct synthetic OOD data from successful trajectories in the
RoboCasa, LIBERO, and RoboTwin~2.0 simulation benchmarks, together with
real-robot AgiBot World data. Task- and phase-aware edits produce
source-aligned counterfactual observations rather than arbitrary visual
variation. Negative edits change a task-relevant execution condition, whereas
robust edits change nuisance appearance while preserving the intended task
state.

\begin{figure}[htbp]
\centering
\includegraphics[width=0.88\columnwidth]{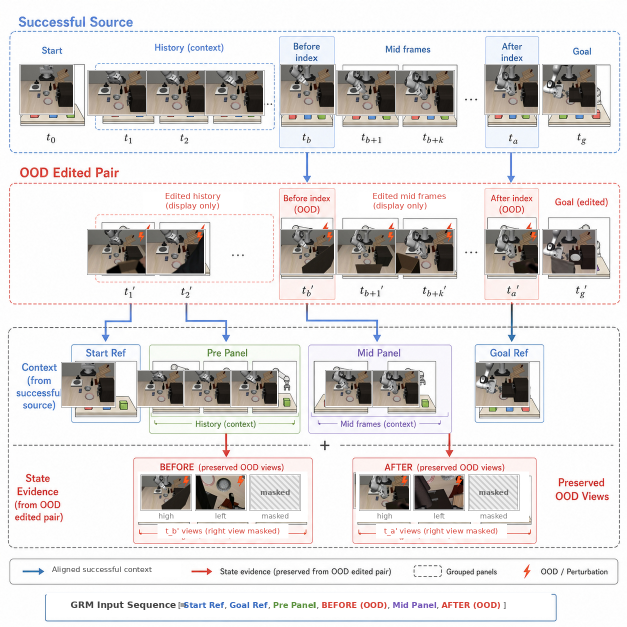}
\caption{OOD provenance case study. All task-reference and temporal-context
images are drawn from the aligned successful source episode, while only the
queried before--after views contain the OOD edits.}
\label{fig:ood_reconstruction_case}
\end{figure}

Figure~\ref{fig:ood_reconstruction_case} illustrates the synthetic-OOD
provenance contract. Task-reference frames and temporal panels are drawn from
the aligned successful source episode, while edited before--after views remain
the actual query evidence. Standard training instead uses same-episode expert
history, and online inference uses the observed rollout history, following the
context contract in the main paper.

Given a successful source trajectory from dataset $d$,
\begin{equation}
\begin{aligned}
    \tau
    &=
    \left\{
    o_i^v
    \,\middle|\,
    i\in\{0,\ldots,T-1\},\;
    v\in\mathcal V_d
    \right\},\\
    \tilde{\tau}_z
    &=
    \left\{
    \tilde{o}_i^v
    \,\middle|\,
    i\in\mathcal S_z(\tau),\;
    v\in\mathcal V_d
    \right\},
\end{aligned}
\label{eq:ood_trajectory_construction}
\end{equation}
where $\mathcal V_d\subseteq\{h,l,r\}$ is the available camera set after
mapping dataset-specific streams to canonical high/third-person, left-wrist,
and right-wrist slots. Unavailable slots are padded and masked by the data
loader. The variable $z$ denotes one OOD operator, and
$\mathcal S_z(\tau)$ is its phase-valid interval. The same operator
specification is applied throughout the interval, using references from the
corresponding source camera.

\paragraph{Pipeline overview.}
The construction proceeds in six steps:
\emph{(i)} normalize the instruction, identify the task-relevant target, and
map the task to manipulation primitives;
\emph{(ii)} assign a semantically compatible negative or robustness operator;
\emph{(iii)} localize the trajectory phase in which the intervention is
meaningful;
\emph{(iv)} construct a multi-reference editing request for every selected
frame and camera;
\emph{(v)} generate the edited observation under operator-specific, geometric,
and view-specific constraints; and
\emph{(vi)} retain source-episode, frame, camera, prompt, and reference
provenance for downstream reward-pair construction.

\subsection{Task Parsing and Operator Assignment}
\label{app:ood_task_parsing}

The task instruction is first normalized and mapped to one or more manipulation
primitives. We use seven primitive categories: \emph{grasp--lift},
\emph{place--transfer}, \emph{open}, \emph{close}, \emph{toggle-on},
\emph{toggle-off}, and \emph{push--move}. A pick-and-place instruction, for
example, activates both grasp--lift and place--transfer, whereas opening a
drawer, toggling a device, or pushing an object activates the corresponding
interaction primitive.

We additionally extract a target phrase from the instruction. The parser
removes the action prefix and truncates the remaining phrase at spatial or
relational markers such as ``into,'' ``onto,'' ``from,'' or ``next to.'' The
resulting phrase is inserted into the generation prompt and used to distinguish
the task object from non-target scene objects.

\paragraph{Primitive-to-operator mapping.}
Grasp--lift tasks admit the negative operators \emph{wrong object},
\emph{empty grasp}, and \emph{target replacement}, together with all robust
operators. Place--transfer tasks admit \emph{wrong object}, \emph{non-release},
and \emph{target replacement}, again together with all robust operators. Open,
close, toggle, and push tasks use the robust operators \emph{foreground
occlusion}, \emph{background change}, \emph{distractor addition or removal},
and \emph{non-target color change}, which preserve the relevant articulated,
on/off, or displacement state.

We apply two semantic guards before assigning an operator. First, for
instructions involving multiple legitimate task objects, identity-changing
edits such as \emph{wrong object} and \emph{target replacement} are avoided
when another valid negative operator is available. This prevents ambiguous
correspondences among multiple admissible targets. Second, when the instruction
explicitly specifies a target color, the \emph{non-target color change}
operator is restricted to non-target objects. Each generated OOD trajectory is
assigned a single operator, so its edited frames form a temporally coherent
counterfactual trajectory rather than a sequence of unrelated perturbations.

\subsection{Phase-Aware Temporal Localization}
\label{app:ood_phase_localization}

Applying a failure edit to an arbitrary frame may produce a visibly different
image without changing the task interpretation. We therefore
localize each interaction-related operator to a primitive-dependent temporal
window. For sampled frame index $i$, normalized source-trajectory progress is
defined as
\begin{equation}
    p_i=\frac{i}{T-1}.
\label{eq:ood_normalized_progress}
\end{equation}
Each frame is assigned an active primitive $\pi_i$. For common composite
instructions, the active primitive changes according to source-trajectory
progress. Grasp-to-place and open-to-place tasks switch to place--transfer at
$p_i=0.45$; place-to-close tasks switch to close at $p_i=0.65$; and
toggle-to-place tasks switch to place--transfer at $p_i=0.35$.

The valid progress intervals are $[0.18,0.80]$ for grasp--lift,
$[0.28,0.95]$ for place--transfer, $[0.15,0.85]$ for open and close,
$[0.10,0.75]$ for toggle-on and toggle-off, and $[0.12,0.85]$ for push--move.
These windows cover the principal interaction phases---approach, contact,
transport, articulation, release, or displacement---while excluding early
frames whose appearance is not yet informative for the selected intervention.

Let
$G_z(\pi,p)\in\{\operatorname{KEEP},\operatorname{EDIT}\}$
denote the fixed operator-specific eligibility rule. For
\emph{non-release}, the rule activates only in the release subphase; the
remaining interaction-related operators activate at their first eligible
task phase. The intervention onset is
\begin{equation}
    i_{\mathrm{start}}
    =
    \min\left\{
    i
    \,\middle|\,
    G_z(\pi_i,p_i)=\operatorname{EDIT}
    \right\}.
\label{eq:ood_start_index}
\end{equation}
The corresponding operator-specific rule is used for \emph{wrong object},
\emph{empty grasp}, \emph{non-release}, \emph{target replacement}, and
\emph{foreground occlusion}. In contrast, \emph{background change},
\emph{distractor addition or removal}, and \emph{non-target color change}
begin at the first sampled state because they represent persistent scene-level
nuisance variation. Applying these operators throughout the trajectory avoids
an artificial appearance discontinuity at the interaction boundary.

For a standard single-stage or pick-and-place trajectory, the intervention is
propagated over the selected suffix:
\begin{equation}
    \mathcal S_z(\tau)
    =
    \left\{
    i_{\mathrm{start}},\ldots,T-1
    \right\}.
\label{eq:ood_selected_interval}
\end{equation}
The active primitive supplied to the prompt may still change from grasping to
transport or release as the trajectory progresses. For explicitly multi-stage
instructions containing distinct subgoals, the selected interval is restricted
to the contiguous segment governed by the primitive active at
$i_{\mathrm{start}}$. This prevents an edit defined for one subgoal from being
propagated into a later subgoal with a different task meaning.

\subsection{Multi-Reference Edit Construction}
\label{app:ood_multireference}

A single queried frame may not reveal the identity or original location of the
target. This occurs when the target is already held, partially occluded by the
gripper, visually similar to nearby objects, or absent from the current field
of view. We therefore condition each edit on an ordered set of observations
from the same successful source trajectory.

For queried frame $i$, the pipeline selects
$K_i\leq13$ reference indices:
\begin{equation}
    \mathcal J_i
    =
    \left\{
    r_1,\ldots,r_{K_i}
    \right\},
    \qquad
    r_k\neq i.
\label{eq:ood_reference_indices}
\end{equation}
The indices are distributed over the complete sampled trajectory and retained
in chronological order. When fewer than 13 alternative frames are available,
all available source frames are used. For camera $v$, the visual input is
\begin{equation}
    \mathcal Q_i^v
    =
    \left[
    o_{r_1}^v,\ldots,o_{r_{K_i}}^v,o_i^v
    \right].
\label{eq:ood_visual_input}
\end{equation}
Thus, each request contains $K_i$ read-only references followed by one editable
image; $K_i=13$ whenever at least 13 alternative frames are available. The queried frame is excluded from the reference set, and the
final image in the sequence is always the only image that may be modified.

All references satisfy three alignment constraints: they originate from the
same successful episode as the queried frame, use exactly the same camera, and
are passed to the editor in temporal order. A head-camera query therefore
receives only head-camera references, while a left- or right-wrist query uses
references from the corresponding wrist camera. This prevents the editor from
transferring the projection, visible object surfaces, or gripper geometry of
one camera to another.

The ordered reference stack serves complementary roles. Early frames provide
the initial target identity, target location, object inventory, and unmodified
scene layout. Frames around the interaction phase reveal how the target
approaches or enters the gripper. Later frames provide evidence about
successful transport, release, or the final articulated state. The references
are used to interpret the task and reconstruct local content; their global
camera pose or complete image layout may not be copied into the queried frame.

Each available camera view is edited through a separate view-specific request:
\begin{equation}
    \tilde{o}_i^v
    =
    \mathcal G\!\left(
    \mathcal Q_i^v;
    \mathcal P(c,\pi_i,z,v)
    \right),
\label{eq:ood_generation}
\end{equation}
where $\mathcal G$ denotes the instruction-following image editor and
$\mathcal P$ denotes the structured prompt. The source index, OOD type, and
task semantics are shared across the three requests, whereas geometric
constraints are specialized to each camera.

\subsection{Structured Prompt Design}
\label{app:ood_prompt_design}

The generation prompt is composed from a fixed set of functional blocks:
\begin{equation}
\begin{aligned}
    \mathcal P
    ={}&
    \mathcal P_{\mathrm{input}}
    \oplus
    \mathcal P_{\mathrm{scene}}
    \oplus
    \mathcal P_{\mathrm{task}}
    \oplus
    \mathcal P_{\mathrm{target}}\\
    &\oplus
    \mathcal P_{\mathrm{operator}}
    \oplus
    \mathcal P_{\mathrm{view}}
    \oplus
    \mathcal P_{\mathrm{physical}}
    \oplus
    \mathcal P_{\mathrm{output}}.
\end{aligned}
\label{eq:ood_prompt_composition}
\end{equation}
This decomposition separates the required intervention from the quantities that
must remain invariant.

\paragraph{Input contract.}
The prompt first states that the request is an image-editing task rather than
text-to-image generation.Images $1$ through $K_i$ are declared read-only references, and image
$K_i+1$ is declared the only editable image. The editor is
prohibited from returning a reference image or reconstructing the scene from a
different camera pose.

The output must preserve the queried resolution of $320{\times}240$, field of
view, crop, perspective, exposure, white balance, color temperature, sharpness,
noise level, depth of field, and compression characteristics. The prompt
explicitly prohibits global enhancement, denoising, restyling, or unnecessary
full-image redrawing.

\paragraph{Scene and target binding.}
The prompt contains the complete task instruction, current frame ID, ordered
reference-frame IDs, and camera identity. It states that all references come
from the same successful trajectory and should be interpreted jointly to
recover the original task process. When a target phrase can be extracted, it is
included explicitly and used together with the trajectory references to
identify the task object. If several visually similar target candidates exist,
a rule that removes, preserves, or moves the target is applied consistently to
all relevant candidates unless the operator is explicitly defined for one
instruction-referenced instance.

\paragraph{Primitive-specific task block.}
A task block specifies whether the current frame belongs to grasping, placement,
opening, closing, toggling, or pushing. The grasp block, for example, identifies
the object originally intended for acquisition, whereas the placement block
emphasizes the transported object, support surface, and target placement region.
This block prevents a visually plausible edit from becoming semantically
inconsistent with the active manipulation stage.

\paragraph{Operator contract.}
The operator block defines the exact counterfactual intervention. For negative
states, it specifies the task-relevant relation to invalidate, such as object occupancy
inside the gripper, target identity, or release completion. For robust states,
it specifies the nuisance factor that may change and explicitly states that the
target, robot state, and task outcome must remain unchanged.

\paragraph{Global physical constraints.}
Across all OOD types, the prompt prohibits moving, rotating, reshaping, or
repositioning the robot arm. Joint configuration, link geometry, and
end-effector pose must remain identical to those in the queried source image.
The gripper position, orientation, and opening are also preserved, except for
the minimal local reconstruction required when removing or inserting held
content.

Every moved, removed, added, replaced, or occluding object must respect the
source perspective, support relation, contact geometry, illumination, shadow
direction, material response, depth ordering, and occlusion pattern. Removed
regions must be completed using surrounding scene texture without duplicated
boundaries, smearing, holes, or inconsistent gripper geometry.

\paragraph{View-specific constraints.}
The prompt contains an additional block specialized to the queried camera.
Head-camera prompts emphasize global scene consistency and the projected grasp
corridor. Wrist-camera prompts restrict editing to the gripper, target, and
immediately adjacent region while preserving visible finger geometry and local
occlusion ordering.

\paragraph{Output contract.}
The editor is instructed to return exactly one image: the edited version of the
final input image. No text, explanation, alternative image, or intermediate
result is permitted. The requested intervention must remain visually
identifiable, but realism and local consistency take priority over exaggerated
change.

In abbreviated form, every prompt specifies
\emph{(i)} which images are read-only references;
\emph{(ii)} which final image is editable;
\emph{(iii)} the instruction, target, and active primitive;
\emph{(iv)} the OOD intervention and its state-dependent branch;
\emph{(v)} camera, robot, object, geometry, lighting, and image-quality
invariants; and
\emph{(vi)} the single-image output requirement.

\subsection{Negative Counterfactual Operators}
\label{app:ood_negative_operators}

Negative operators change a task-relevant execution condition while keeping the
source robot pose and trajectory phase fixed.

\paragraph{Wrong object.}
If the gripper is empty or approaching the target, the task target in the grasp
corridor is replaced by a visually distinct non-target object, while the
original target is relocated to a plausible position outside the corridor.
This makes the impending interaction correspond to the wrong object rather
than an empty grasp. If the gripper is occupied, only the held content is
replaced by an object with a different category, dominant color, and
silhouette. The original target remains naturally situated elsewhere whenever
visible.

\paragraph{Empty grasp.}
The defining condition is an empty region between the gripper fingers. If the
gripper is approaching an object, the target and any object intersecting the
opening, closing region, or forward grasp corridor are moved laterally outside
the grasp path. If the gripper already contains an object, that object is
removed without changing the gripper pose or opening. Only the minimal local
region is reconstructed to recover the visible fingers, interior background,
shadows, and reflections. The target is returned to its source-scene location
when this location is supported by the trajectory references.

\paragraph{Non-release.}
The \emph{non-release} operator converts a successful release or near-release
state into a continued-holding state. If the target has already left the
gripper, the same target is inserted back into the current gripper and its
released instance is removed to avoid duplication. The gripper pose is not
replaced by an earlier source pose. If the target is already held, its identity
and appearance are preserved while the continued-holding configuration remains
unchanged.

\paragraph{Target replacement.}
Only the instruction-referenced target is replaced. The replacement preserves
the original target's approximate image footprint, location, local support or
grasp relation, and occlusion order, but differs clearly in category and
silhouette. All surrounding objects, articulated states, and robot geometry
remain fixed. The resulting state therefore preserves the local spatial
arrangement while violating task-object identity. Unlike \emph{wrong object}, this operator changes the identity of the
instruction-referenced scene target rather than the identity of the object
currently approached or held by the gripper.

\subsection{Robustness-Preserving Operators}
\label{app:ood_robust_operators}

Robustness operators change image appearance without changing task semantics and
therefore retain the same progress value as their aligned source states.

\paragraph{Foreground occlusion.}
A plausible close-range occluder, such as a hand, board, box, or cloth region,
is inserted in front of the critical interaction area. The occluder may
partially hide the target or gripper, but it may not cover the complete image.
The underlying robot--object state, target identity, contact state, task
outcome, and all pre-existing scene objects remain unchanged. The occluder is
placed according to the queried camera's depth ordering rather than used to
redraw the hidden interaction into a different state.

\paragraph{Background change.}
One or two small modifications are applied to secondary background regions,
such as a minor layout adjustment or local appearance variation away from the
interaction area. Target identity and location, robot and gripper state,
articulated-object state, and the geometry of the critical contact region remain
unchanged.

\paragraph{Distractor addition or removal.}
A small number of non-target objects are added to or removed from background or
secondary regions. Added objects must have plausible scale, support, material,
illumination, and shadow; removed regions are completed using neighboring scene
texture. New objects are not placed adjacent to the gripper or target, and no
task-relevant object or receptacle may be removed.

\paragraph{Non-target color change.}
A small, natural change in color, saturation, or brightness is applied to one or
two non-target objects. No object is added, removed, moved, rotated, or
replaced. The target color and identity, object geometry, scene layout, robot
state, contact relations, and task outcome remain fixed. When color is part of
the instruction, the target color is protected by both operator selection and
the prompt.

\subsection{Camera-Specific Geometric Constraints}
\label{app:ood_camera_constraints}

The same semantic edit requires different geometric instructions under the head
and wrist cameras. We therefore append a camera-specific block to each prompt.

\paragraph{Head camera.}
The head view exposes the global scene layout and the projected grasp corridor.
For \emph{empty grasp}, the prompt gives highest priority to clearing the
region between the fingers, directly below the gripper, and along its projected
approach path. For \emph{wrong object}, it verifies that no valid target
remains within the visually reachable region around an empty gripper. For
\emph{non-release}, it verifies that the final edited gripper is visibly
non-empty.

\paragraph{Wrist cameras.}
Wrist views are treated as local close-up observations. Editing is restricted
to the gripper, target, and immediately adjacent image region. The prompt
prohibits expanding an object surface that is occluded in the queried wrist
view, copying geometry from the head camera, or producing the opposite wrist's
observation. The left-wrist prompt preserves the visible left finger, contact
boundary, and local reflection pattern; the right-wrist prompt applies the
analogous constraint to the visible right-side geometry.

When an object is inserted, removed, or replaced inside the gripper, its
boundary must remain behind the visible metal finger edges according to source
depth ordering. If the robot itself occludes the interaction region, that
occlusion is preserved rather than removed merely to make the edit more
explicit. These constraints encourage the available edited views to express the same
trajectory-level OOD semantics while respecting their distinct projections
and visible local geometry.

\subsection{Generation and Provenance}
\label{app:ood_generation_provenance}

The OOD construction uses Gemini 3.1 Flash Image as the
instruction-following image-editing backend in batch mode. Each
frame--camera pair is submitted as one editing request containing the ordered
reference images, the final queried image, and the fully instantiated prompt.

Every request receives a unique identifier. Its record contains the source dataset, task instruction, episode ID, frame ID,
normalized frame index, camera, active primitive, OOD type, OOD group,
reference-frame IDs, generation prompt, backend identifier, generation
timestamp, and source path. The identifier is subsequently used to associate each returned
image with the exact source frame and camera. At the trajectory level, a
manifest records the selected source-frame interval, operator, OOD group, onset
mode, primary primitive, multi-stage indicator, multi-object indicator,
reference configuration, and complete list of edited source-frame IDs.

This metadata preserves the alignment
\begin{equation}
    \tilde{o}_i^v
    \longleftrightarrow
    o_i^v
    \longleftrightarrow
    (\tau,i,v,z).
\label{eq:ood_provenance_alignment}
\end{equation}
The downstream loader can therefore reconstruct both the original successful
observation and its edited counterpart at the same source progress index:
\begin{equation}
\begin{aligned}
    X_i
    &=
    \left\{
    o_i^v
    \mid v\in\mathcal V_d
    \right\},\\
    \tilde{X}_i^z
    &=
    \left\{
    \tilde{o}_i^v
    \mid v\in\mathcal V_d
    \right\}.
\end{aligned}
\label{eq:ood_multiview_state}
\end{equation}

For every OOD pair, all task-reference and temporal-context images are
reconstructed exclusively from the aligned successful source episode. The
generated trajectory contributes only the queried before--after observations.
This source-only context contract preserves the source phase and panel ordering
without replacing the wrong-object, empty-grasp, non-release,
target-replacement, occlusion, or background-perturbation evidence contained
in the OOD query.

The recorded provenance keeps the successful source state, edited observation,
camera projection, and temporal context separately traceable. Negative and
robust states are therefore compiled into the signed progress space using their
shared source index rather than independently sampled trajectories.

\section{Computing Infrastructure and Training Cost}
\label{app:compute}

Reward-model fine-tuning and RoboTwin simulation policy learning use eight
NVIDIA H100 GPUs with 80\,GB of memory per GPU (640\,GB aggregate accelerator
memory) for every training run. Real-world ConRFT training instead uses one
NVIDIA RTX~4090. The retained experiment records do not provide verified host
CPU, system RAM, operating-system, or CUDA details, so we report only the
accelerator allocations rather than inferring unverified model or version
information.
A complete reward-model training run takes approximately 20 hours in this
eight-GPU setting. Reward-model training uses 400K signed pairs across two
200K stages, together with the fixed QA200K auxiliary mix excluded from this
pairwise budget. Each trainable reward-model configuration and policy-learning
condition is trained once, and the GRM remains frozen in all downstream RL
experiments.

\section{Simulation Reinforcement Learning Details}
\label{app:simulation_rl}

We optimize task-specific OpenVLA-OFT SFT checkpoints with GRPO in RLinf on
the RoboTwin~2.0 tasks \texttt{place\_empty\_cup},
\texttt{place\_container\_plate}, and \texttt{handover\_block}. The GRM
remains frozen, and history-enabled conditions construct panels from the
observed rollout prefix. The policy receives head-camera RGB and 14-D
proprioception, predicts 14-D actions in 25-action chunks, and is LoRA-tuned
in BF16 with FSDP. GRPO uses 128 parallel environments, groups of eight
rollouts, normalized group-relative advantages, and $\gamma=1$.

The sparse baseline is trained for 1,000 epochs and C00--C11 for 300 epochs. We
run deterministic 128-episode evaluations every 20 epochs over fixed reset
states. For \texttt{place\_empty\_cup}, \texttt{place\_container\_plate}, and
\texttt{handover\_block}, respectively, the episode horizons are 200, 150, and
400 steps and the learning rates are $10^{-4}$, $2{\times}10^{-4}$, and
$2{\times}10^{-4}$.

\section{Real-World Experiments}
\label{app:real_world}

\subsection{Platform, Observations, and Task}
\label{app:real_world_platform}

The physical workcell contains two Franka manipulators, a Pika teleoperation
interface, and calibrated ZED cameras. The camera system supplies synchronized
wrist and third-person RGB observations to both the reward model and the
policy. The GRM remains frozen during online policy learning. History-enabled
conditions construct temporal panels from the observed rollout prefix; no
successful source alignment or repetition label is provided online.

\begin{figure}[!t]
\centering
\includegraphics[width=\columnwidth]
{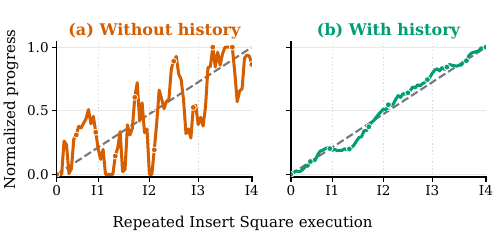}
\caption{Logged progress predictions from matched static and
rollout-history-conditioned executions of repeated \emph{Insert Square}.
Tick $I_k$, $k\in\{1,\ldots,4\}$, denotes the $k$-th insertion attempt, and
the gray dashed line denotes execution order. Static predictions exhibit
non-monotonic jumps at recurring states, whereas rollout-history conditioning
produces more order-consistent estimates.}
\label{fig:appendix_history_progress_case}
\end{figure}

We evaluate the single-arm \emph{Insert Square} task. The acting arm grasps a
square block containing four holes, adjusts its position and orientation, and
inserts it onto four upright pegs on the target board. Successful insertion
requires millimeter-scale alignment. To expose temporal aliasing rather than
only one-shot manipulation accuracy, each trial contains $K=4$ consecutive
back-and-forth insertion attempts. Consequently, similar pickup, alignment, and insertion observations recur at
different cumulative progress levels.
Figure~\ref{fig:appendix_history_progress_case} compares representative static
and rollout-history-conditioned progress traces.

\subsection{Offline-to-Online ConRFT Protocol}
\label{app:real_world_conrft}

We use ConRFT as the policy-optimization procedure. Its
policy retains a frozen Octo-Small observation backbone and replaces the
original action head with a consistency policy; a learned critic supplies the
Q objective. Offline initialization uses 40 teleoperated trajectories in a
demonstration buffer $\mathcal D$. The policy is optimized with behavior
cloning and calibrated Q-learning, while the critic includes a conservative
penalty to stabilize value estimates outside the demonstration distribution.

During online adaptation, autonomous transitions are appended to replay buffer
$\mathcal R$. A human operator intervenes when an action is unsafe or likely to
cause failure, and corrected trajectories are returned to demonstration buffer
$\mathcal D$. Updates sample equally from $\mathcal D$ and $\mathcal R$.
The online policy retains the behavior-cloning and Q terms with weights shifted
toward reward-driven improvement, while the critic uses the Bellman error
without the offline conservative penalty. Intervention and data-routing rules
are identical across reward conditions.

Table~\ref{tab:appendix_real_conrft_hparams} summarizes the real-world
optimization settings.

\begin{table}[!t]
\centering
{\small
\setlength{\tabcolsep}{4.0pt}
\begin{tabular}{@{}lcc@{}}
\toprule
Hyperparameter & Symbol & Value \\
\midrule
Global batch size & -- & 256 \\
Learning rate & -- & $3{\times}10^{-4}$ \\
Reward discount & $\gamma$ & 0.98 \\
Demonstrations & -- & 40 \\
Insertion attempts / trial & $K$ & 4 \\
Online episodes (C00--C11) & -- & 200 \\
Online episodes (Sparse) & -- & 300 \\
Evaluation trials / condition & -- & 20 \\
Insertion attempts / condition & -- & 80 \\
Offline BC weight & $\beta$ & 1.0 \\
Offline Q weight & $\eta$ & 0.1 \\
Conservative penalty & $\alpha$ & 0.1 \\
Online BC weight & $\beta'$ & 0.5 \\
Online Q weight & $\eta'$ & 1.0 \\
Compute & -- & $1{\times}$ RTX 4090 \\
\bottomrule
\end{tabular}
}
\caption{Optimization settings for the real-world ConRFT experiments.}
\label{tab:appendix_real_conrft_hparams}
\end{table}

\subsection{Controlled Comparison and Metrics}
\label{app:real_world_evaluation}

Every reported condition uses the same event-sparse task reward, emitted once
after each successful insertion. The Event Sparse baseline uses only this
reward, whereas C00--C11 add GRM shaping. A terminal-only pilot rewarding only
completion of all four attempts did not learn and is excluded from the reported
comparison. All reported variants start from the same policy initialization and
use identical demonstration data, observations, resets, interventions, and
evaluation. C00--C11 use 200 online episodes and Event Sparse uses 300. Each condition is then evaluated in 20 trials under the shared evaluation
protocol.
This design isolates the effects of OOD supervision and ordered history while
making the larger sparse-reward interaction budget explicit. We report the
number of successful insertions among the resulting 80 attempts and the number
of full-sequence successes among 20 trials. A trial contributes to the latter
count only when all four insertion attempts succeed.

\section{Additional Ablation Details}
\label{app:ablation}

\subsection{Additional Signed-Hop Diagnostics}
\label{sec:appendix_signed_hop}

The main paper reports the curriculum controls and the per-family improvement
of the selected schedule. Here we separate the Stage~2 movement for the two
large-to-fine variants and provide the complete replay-rate sensitivity sweep.
The first 200K large-Hop examples are shared by E3 and E4; their second stages
differ only in whether large-Hop pairs are replayed.

\begin{table}[t]
\centering
{\small
\setlength{\tabcolsep}{2.3pt}
\begin{tabular}{@{}llcccc@{}}
\toprule
Strategy & Step & Temp. & Neg. & Rob. & OOD-T \\
\midrule
No replay & 200K & 0.9860 & 0.9945 & 0.9510 & 0.9870 \\
No replay & 400K & 0.9950 & 0.9937 & 0.9606 & 0.9924 \\
25\% replay & 200K & 0.9860 & 0.9945 & 0.9510 & 0.9870 \\
25\% replay & 400K & \textbf{0.9953} & \textbf{0.9945}
& \textbf{0.9618} & \textbf{0.9931} \\
\bottomrule
\end{tabular}
}
\caption{Stage~2 VOC for the two large-to-fine curricula. Moderate replay
yields slightly higher final scores on the reported temporal and OOD
families.}
\label{tab:stage2_ablation}
\end{table}

\begin{figure}[t]
\centering
\includegraphics[width=\columnwidth]
{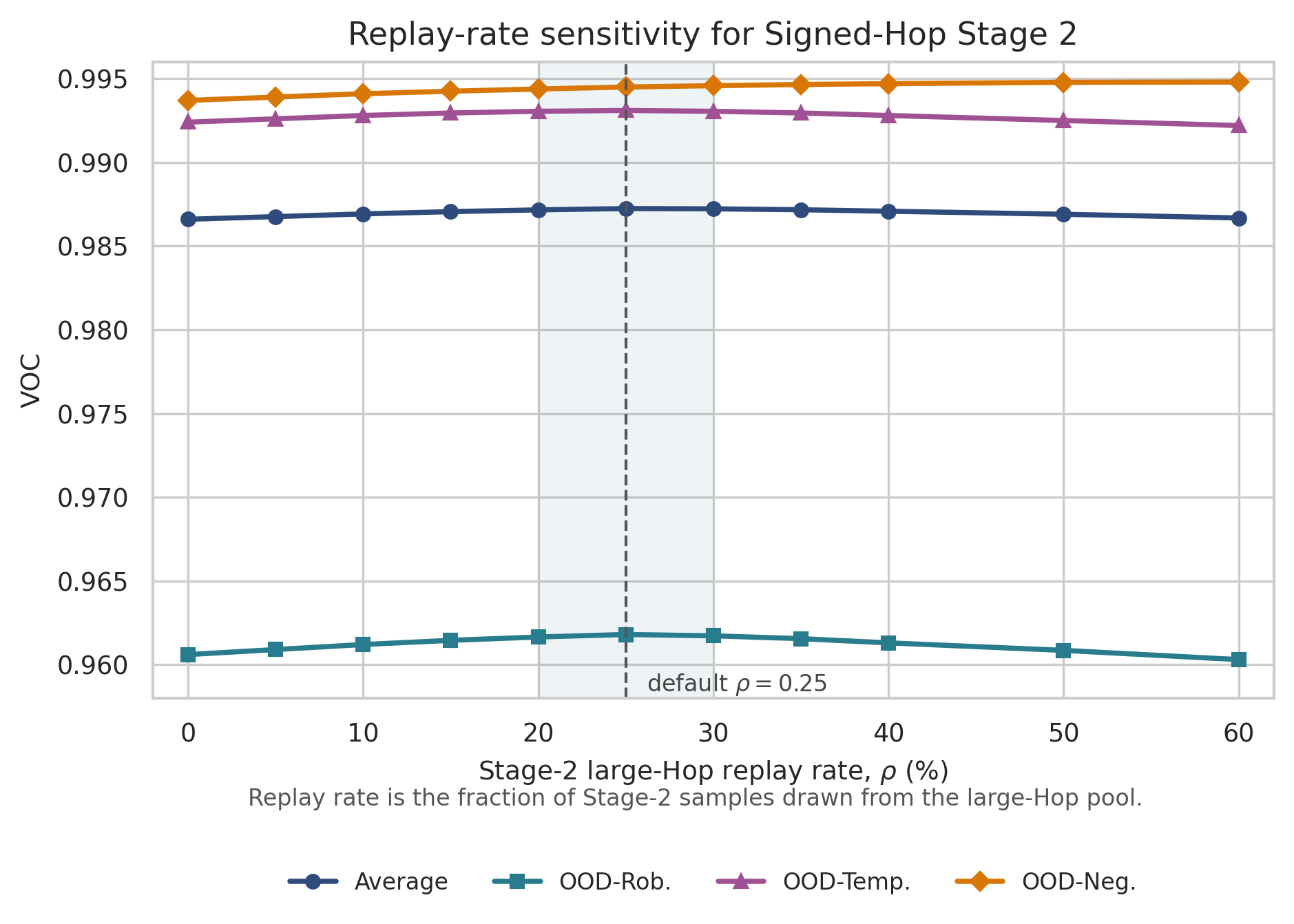}
\caption{Sensitivity to the fraction $\rho$ of Stage~2 samples replayed from
the large-Hop pool. The shaded neighborhood marks the moderate-replay regime
around the selected $\rho=0.25$ setting.}
\label{fig:appendix_replay_sensitivity}
\end{figure}

The replay rate changes only the Stage~2 mixture; it does not redefine the
large-Hop pool used in Stage~1. Average VOC increases from 0.9866 without
replay to 0.98724 at $\rho=0.25$, remains nearly unchanged at $\rho=0.30$
(0.98722), and decreases to 0.98668 at $\rho=0.60$
(Figure~\ref{fig:appendix_replay_sensitivity}). The shallow maximum near
$\rho=0.25$ is consistent with moderate replay retaining large-transition
examples while preserving most of the fine- and zero-Hop calibration budget.

OOD-negative VOC increases from 0.9937 at $\rho=0$ to 0.9948 at
$\rho=0.60$, whereas OOD-robust and OOD-temporal VOC peak near moderate replay
and decline at larger ratios. The selected $\rho=0.25$ achieves the highest
observed overall average rather than the best score on every branch. Replay is
used only during reward-model training and is unrelated to inference-time
memory or the RL replay buffer.

\FloatBarrier

\subsection{Trajectory-Level Case Studies}
\label{sec:appendix_trajectory_cases}

\begin{figure}[t]
\centering
\includegraphics[width=\columnwidth]
{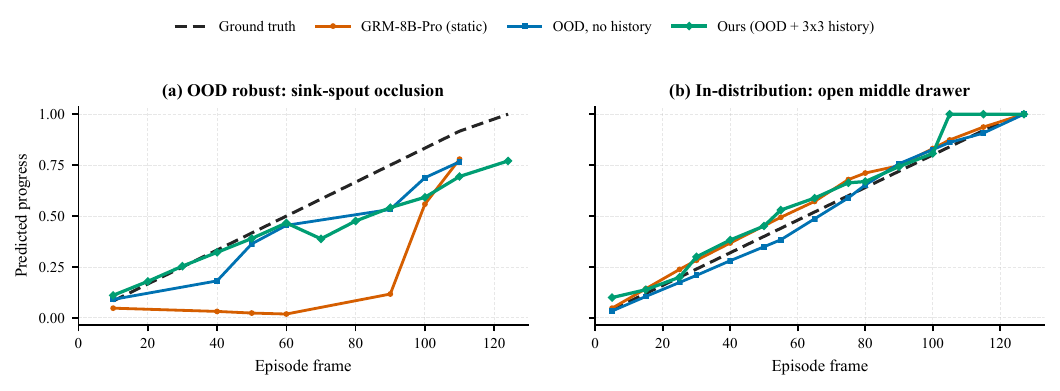}

\medskip

\includegraphics[width=0.47\columnwidth]
{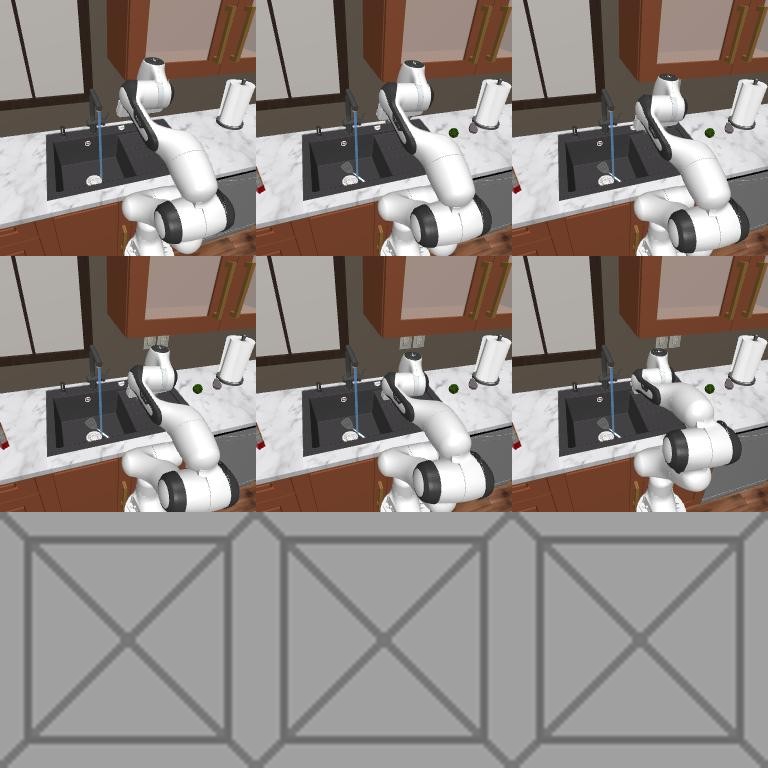}
\hfill
\includegraphics[width=0.47\columnwidth]
{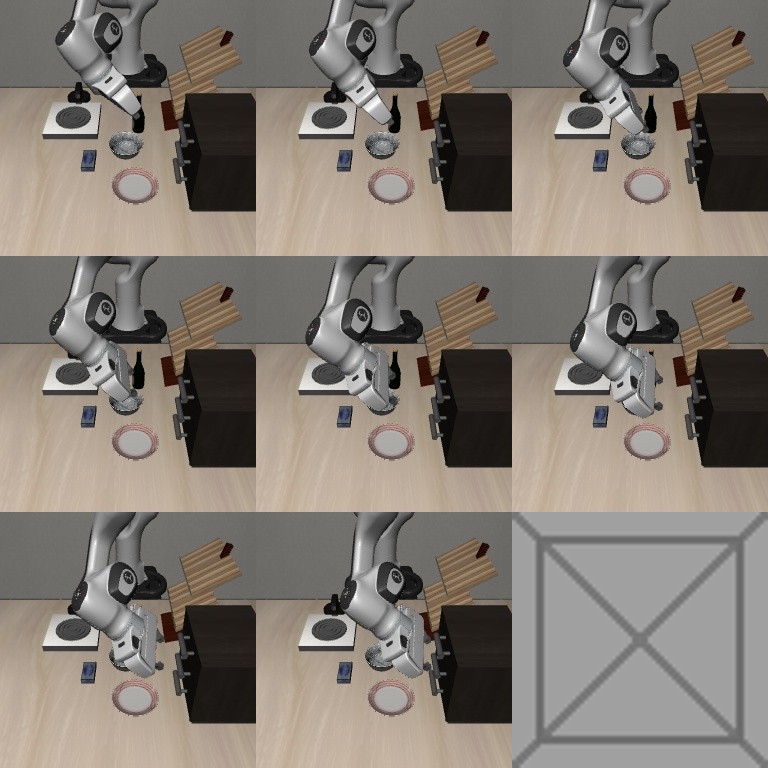}

\caption{Episode-level progress traces on two held-out cases. Top: progress
curves for a RoboCasa sink-spout episode with foreground occlusion and an
in-distribution LIBERO middle-drawer episode. Bottom left: the source-reference
pre-query panel for the sink-spout case. Bottom right: the same-episode
pre-query panel for the drawer case. Gray cells denote fixed padding.}
\label{fig:appendix_episode_curves}
\end{figure}

Figure~\ref{fig:appendix_episode_curves} complements the aggregate VOC results
with representative trajectory-level progress curves.

On the foreground-occlusion case, the static GRM remains close to zero through
the middle of the episode and then jumps late. OOD supervision without history
recovers much of the trend, while the panel-conditioned model follows the
reference ordering most closely in the early and middle phases. The remaining
late-episode gap shows that history does not eliminate all sensitivity to a
strong foreground edit.

The in-distribution drawer case serves as a calibration check rather than an
OOD stress test. All three learned curves preserve the episode order and reach
completion. The same-episode panel-conditioned curve follows the reference
closely over most of the rollout but saturates early at the final evaluated
states. We retain this behavior rather than smoothing or selecting only
favorable points, exposing both the benefit of temporal context and the
remaining endpoint-calibration error.

\FloatBarrier

\subsection{Compact Ablation Tables}
\label{sec:appendix_compact_ablations}

Table~\ref{tab:appendix_ablation_grid} reports the complete checkpoint and
configuration comparisons underlying the main ablation summary.

\par\smallskip
\noindent
\begin{minipage}{\columnwidth}
\centering
{\small
\setlength{\tabcolsep}{1.8pt}
\begin{tabular}{@{}llcccc@{}}
\toprule
Base & Cond. & Type & Stat. & Hist. & Gain \\
\midrule
GRM & base & Obs. & 0.958 & 0.965 & +0.007 \\
Robo-Dopamine 2.0 & $3{\times}3$+QA & Obs. & 0.967 & 0.986 & +0.019 \\
\midrule
GRM & no OOD & Swp. & 0.954 & 0.962 & +0.008 \\
GRM & OOD+QA & Swp. & 0.967 & 0.977 & +0.010 \\
GRM & OOD+$2{\times}2$ & Swp. & 0.957 & 0.985 & +0.028 \\
GRM & OOD+$3{\times}3$ & Swp. & 0.955 & 0.988 & +0.033 \\
GRM & OOD+$3{\times}3$+QA & Swp. & 0.958 & 0.988 & +0.030 \\
Robo-Dopamine 2.0 & OOD & Swp. & 0.965 & 0.974 & +0.009 \\
Robo-Dopamine 2.0 & OOD+QA & Swp. & 0.967 & 0.977 & +0.010 \\
Robo-Dopamine 2.0 & OOD+$2{\times}2$ & Swp. & 0.962 & 0.985 & +0.023 \\
Robo-Dopamine 2.0 & OOD+$3{\times}3$ & Swp. & 0.960 & 0.987 & +0.027 \\
Robo-Dopamine 2.0 & OOD+$3{\times}3$+QA & Swp. & 0.967 & 0.986 & +0.019 \\
\bottomrule
\end{tabular}
}
\captionof{table}{Compact ablation grid supporting the main summary. Obs.
denotes the main reported checkpoint comparison, and Swp. denotes the
controlled configuration sweep. Stat. and Hist. denote \texttt{static8} and
\texttt{history\_panel} VOC, respectively.}
\label{tab:appendix_ablation_grid}
\end{minipage}
\par\smallskip

The controlled sweep shows complementary effects from OOD supervision and
structured temporal panels. The QA mix provides a smaller additional effect
once history panels are enabled.

Table~\ref{tab:appendix_layout_sweep} further examines the interaction between
panel resolution and the number of between-state panels.

\par\smallskip
\noindent
\begin{minipage}{\columnwidth}
\centering
{\small
\setlength{\tabcolsep}{0.7pt}
\renewcommand{\arraystretch}{1.03}
\begin{tabular}{@{}lcccccc@{}}
\toprule
Grid & $B{=}0$ & $B{=}1$ & $B{=}2$
& $B{=}3$ & $B{=}4$ & $B{=}5$ \\
\midrule
\multicolumn{7}{l}{\textit{GRM}} \\
$1{\times}1$ & 0.965 & 0.971 & 0.969 & 0.967 & 0.963 & 0.959 \\
$2{\times}2$ & 0.980 & 0.986 & 0.984 & 0.981 & 0.978 & 0.974 \\
$3{\times}3$ & 0.988 & \textbf{0.992} & \textbf{0.991}
& 0.989 & 0.986 & 0.982 \\
$4{\times}4$ & 0.981 & 0.987 & 0.985 & 0.983 & 0.979 & 0.975 \\
\midrule
\multicolumn{7}{l}{\textit{Robo-Dopamine 2.0}} \\
$1{\times}1$ & 0.966 & 0.972 & 0.970 & 0.967 & 0.964 & 0.960 \\
$2{\times}2$ & 0.981 & 0.987 & 0.985 & 0.982 & 0.979 & 0.975 \\
$3{\times}3$ & 0.991 & \textbf{0.993} & \textbf{0.992}
& 0.992 & 0.991 & 0.988 \\
$4{\times}4$ & 0.982 & 0.988 & 0.986 & 0.984 & 0.980 & 0.976 \\
\bottomrule
\end{tabular}
}
\captionof{table}{Memory-layout sweep. $B$ denotes the number of between-state
panels. Both backbones perform best with $3{\times}3$ panels and one or two
between-state panels.}
\label{tab:appendix_layout_sweep}
\end{minipage}
\par\smallskip

Increasing panel capacity is not uniformly beneficial. Both backbones favor
$3{\times}3$ panels with one or two between-state panels, whereas larger grids
or panel counts provide no consistent improvement.

\begin{figure}[t]
\centering
\includegraphics[width=\columnwidth]
{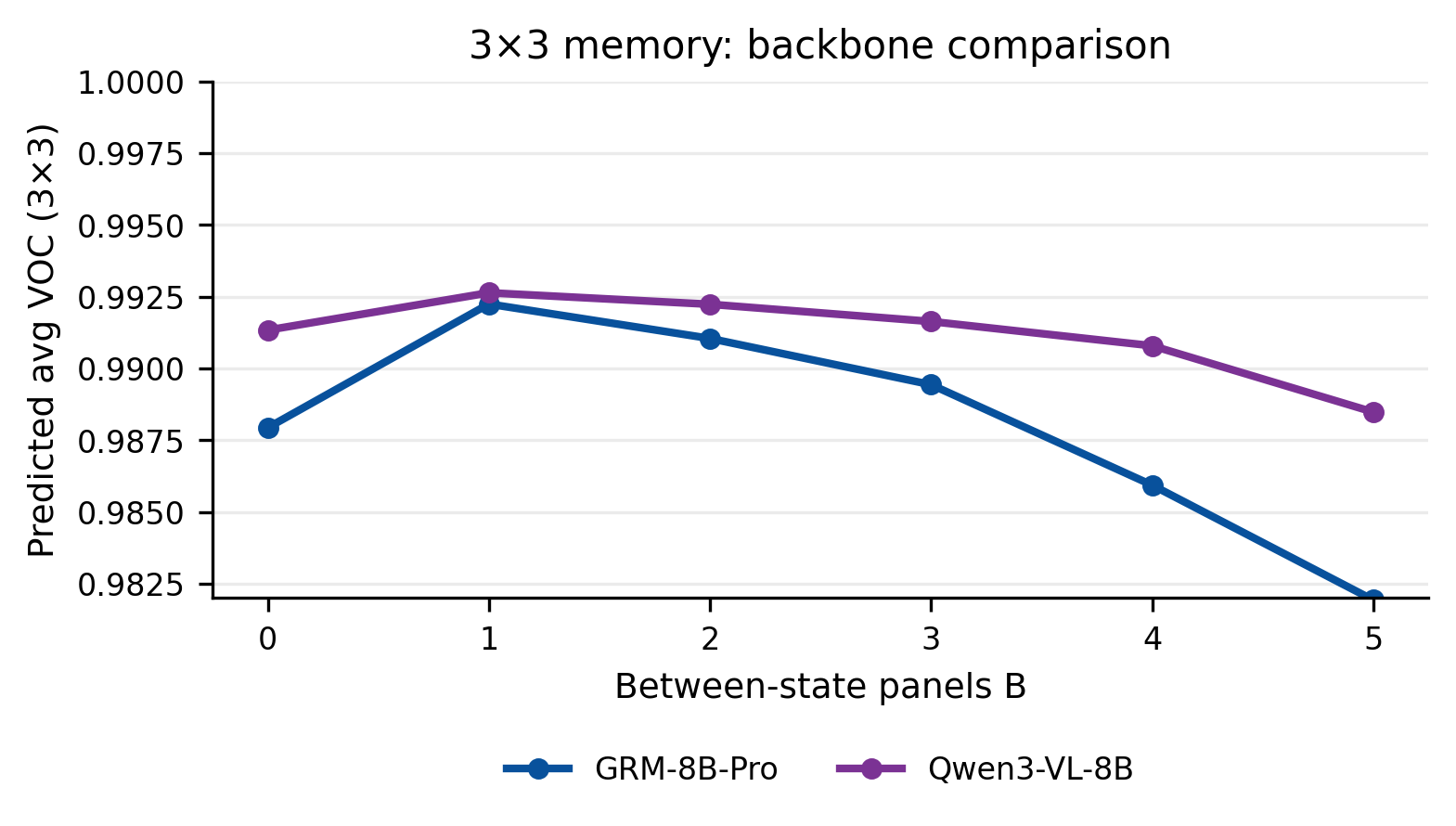}
\caption{$3{\times}3$ memory-panel sweep over between-state panel count.
Both backbones peak with one or two between-state panels; additional panels
do not monotonically improve VOC.}
\label{fig:backbone_3x3}
\end{figure}

Figure~\ref{fig:backbone_3x3} visualizes the same panel-count trend for the
$3{\times}3$ configuration.

The curve confirms that the main benefit comes from adding a limited amount of
between-state evidence rather than continuously increasing the visual-token
load. Runtime panel construction is retained as a diagnostic because it also
tests data-loading-time assembly and prompt alignment.

Overall, the ablations consistently favor structured $3{\times}3$ panels with
one or two between-state panels. OOD supervision supplies the main robustness
gain, while QA mixing and additional panel capacity have smaller effects once
structured temporal context is available.

\end{document}